\documentclass[]{aastex63}
\usepackage{graphicx} 
\usepackage{tcolorbox}
\usepackage{pdfpages}
\usepackage[frozencache,cachedir=.]{minted}

\makeatletter
\AtBeginDocument{\let\LS@rot\@undefined}
\makeatother

\tcbuselibrary{minted,breakable,xparse,skins}
\definecolor{bg}{gray}{0.95}

\DeclareTCBListing{mintedbox}{O{}m!O{}}{%
  breakable=true,
  listing engine=minted,
  listing only,
  minted language=#2,
  minted style=default,
  minted options={%
    linenos,
    gobble=0,
    breaklines=true,
    breakafter=,,
    fontsize=\small,
    numbersep=8pt
    #1},
  boxsep=0pt,
  left skip=0pt,
  right skip=0pt,
  left=25pt,
  right=0pt,
  top=3pt,
  bottom=3pt,
  arc=5pt,
  leftrule=0pt,
  rightrule=0pt,
  bottomrule=2pt,
  toprule=2pt,
  colback=bg,
  colframe=orange!70,
  enhanced,
  overlay={%
    \begin{tcbclipinterior}
    \fill[orange!20!white] (frame.south west) rectangle ([xshift=20pt]frame.north west);
    \end{tcbclipinterior}},
  #3}

\shortauthors{Ali-Dib \& Menou}

\begin{document}

\title{Physics simulation capabilities of LLMs}
\author[0000-0002-6633-376X]{Mohamad Ali-Dib}\footnote{malidib@nyu.edu}
\affiliation{ Center for Astrophysics and Space Science (CASS), New York University Abu Dhabi, PO Box 129188, UAE}
\author{Kristen Menou}\footnote{kristen.menou@utoronto.ca}
\affiliation{Physics \& Astrophysics Group, DPES,  University  of  Toronto  Scarborough, Ontario, M1C 1A4, Canada}
\affiliation{David A. Dunlap Department  of Astronomy \& Astrophysics, University of Toronto, Ontario, M5S 3H4, Canada}
\affiliation{Department of Physics, University of Toronto, Ontario, M5S 1A7, Canada}
\affiliation{Observatoire de la Cote d'Azur, TOP team, Laboratoire Lagrange - CNRS, Nice, France}

\begin{abstract}
 Large Language Models (LLMs) can solve some undergraduate-level to graduate-level physics textbook problems and are proficient at coding. Combining these two capabilities could one day enable AI systems to simulate and predict the physical world.

We present an evaluation of state-of-the-art (SOTA) LLMs on PhD-level to research-level computational physics problems.  We condition LLM generation on the use of well-documented and widely-used packages to elicit coding capabilities in the physics and astrophysics domains. We contribute $\sim 50$ original and challenging problems in celestial mechanics (with {\tt \texttt{REBOUND}}), stellar physics (with {\tt MESA}), 1D fluid dynamics (with {\tt Dedalus}) and non-linear dynamics (with {\tt SciPy}). 
Since our problems do not admit unique solutions, we evaluate LLM performance on several soft metrics:  counts of lines that contain different types of errors (coding, physics, necessity and sufficiency) as well as a more ‘educational’ Pass-Fail metric focused on capturing the salient physical ingredients of the problem at hand. 

As expected, today’s SOTA LLM (GPT4) zero-shot fails most of our problems, although about 40\% of the solutions could plausibly get a passing grade. About $70-90 \%$ of the code lines produced are necessary, sufficient and correct (coding \& physics). Physics and coding errors are the most common, with some unnecessary or insufficient lines. We observe significant variations across problem class and difficulty. 
We identify several failure modes of \texttt{GPT4} in the computational physics domain, such as poor physical units handling, poor code versioning, tendency to hallucinate plausible sub-modules,  lack of physical justification for global run parameters (e.g., simulation time, or upper-lower bounds for parametric exploration) and inability to define steady-state or stopping conditions reliably.

Our reconnaissance work provides a snapshot of current computational capabilities in classical physics and points to obvious improvement targets if AI systems are ever to reach a basic level of autonomy in physics simulation capabilities. 
\end{abstract}

\section{Introduction}

{Large language models (LLMs) are a class of artificial intelligence models that leverage vast amounts of textual data to learn and generate human-like text, enabling them to perform a wide range of natural language processing tasks with impressive performance \citep{llmreview1,llmreview2}.}

LLMs exhibit a wide array of capabilities, including in science-related fields \citep{kaddour2023challenges}. Top-performing LLMs have strong coding abilities \citep[e.g.,][]{2023arXiv231102640K, liu2023mlbench} and their capabilities in biology, chemistry, math and physics are being actively explored \citep[e.g.,][]{wang2023scibench, 2023arXiv231104929B,2023arXiv230913059A,2023arXiv231003731W}. The majority of LLM science evaluations to date has focused on high school to undergraduate textbook cases, partly because it matches the performance level of today’s SOTA LLMs. At the same time, the performance of LLMs on standard benchmarks shows evidence of saturation \citep{2023arXiv231003715M}, which suggests the need for more challenging evaluation methods geared towards next generation models. Indeed, stronger-capability benchmarks and evaluations have recently started to emerge \citep[e.g.][]{rein2023gpqa, mialon2023gaia, 2023arXiv231107361R}.

The pace of progress in LLM capabilities over the
past few years has been remarkable, with steady breakthroughs through scaling, from GPT 2 in 2019 to GPT 4 in 2023 \citep[e.g.][]{bowman2023things}, while multimodal capabilities are now just becoming available \citep[e.g.,][]{GPT4V_card, yang2023dawn}. This rapid progress makes it possible to contemplate a future where LLMs, or related AI systems, 
perform at the level of broadly educated humans \citep[e.g.,][]{morris2023levels,Anthropic_core} and eventually at the level of
scientific research assistants \citep{GPT2030,liu2023mlbench}. The possibility
that AI systems advance further to a level where they start producing original scientific solutions (that would presumably be validated by human experts) has also been considered in the literature, largely in relation to the exceptional alignment and safety challenges that such a situation would create \citep[e.g.][]{ELK, hubinger2023conditioning,michael2023debate,browncohen2023scalable}.
Of potential relevance, a recent proof-of-principle that superhuman AI-capabilities can be learned from by human experts has been presented in the specific chess domain \citep{schut2023bridging}. Taking a broader view, the evaluation of science capabilities of LLMs and related AI systems could eventually have implications for the pursuit of science and for safety and alignment practices.

In this paper, we introduce a specific approach to the evaluation of the physics capabilities of LLMs at the graduate/research level. Our main focus is on the ability of LLMs to reliably generate code to simulate complex physical scenarios of interest in academic research, at a level that is routinely performed by graduate students and research scientists. By construction, our evaluation framework approaches the limit of domain expertise in a given academic discipline. We anticipate that some of the basic elements of our approach could transfer to other simulation-rich disciplines.

Numerical simulations are a work-horse of scientific research in various sub-fields of physics. They require a combination of specialized skills in coding and in physics. In the past decade or so, the incremental availability of flexible and reliable open-source packages built to solve general classes of scientific problems in a given sub-field has greatly improved the reliability and reproducibility of simulation work using such standard packages. Building on this practice, we make use of well-documented and thoroughly-tested open-source simulation tools in our work: {\tt \texttt{REBOUND}}\footnote{https://rebound.readthedocs.io/en/latest/} for celestial mechanics \citep{2012A&A...537A.128R,2020MNRAS.491.2885T}, {\tt MESA}\footnote{https://docs.mesastar.org/} for stellar physics \citep{2011ApJS..192....3P}, {\tt Dedalus}\footnote{https://dedalus-project.readthedocs.io/en/latest/} for
fluid/continuum physics \citep{2020PhRvR...2b3068B} and {\tt SciPy}\footnote{https://docs.scipy.org/doc/scipy/} for non-linear dynamics \citep{2020NatMe..17..261V}.

The use of these open-source tools promotes transparency
in our own analysis but it also serves a specific design
purpose. The wide availability and demonstrated uses of these tools online 
makes it more likely that an LLM has seen numerous relevant code
examples during pre-training to be able to generalize using these tools on 
new problems. Given the strong coding abilities of LLMs, but their (currently) limited abilities in physics, this approach allows us to anchor the generation task in our problems on a specific code base, potentially surfacing shortcomings in the physics world models learned by LLMs.

Another reason to focus on physics simulations (or scientific simulations
in general) is that simulation capabilities could eventually become a factor in the release of LLM agents interacting with the world. Indeed, an ability to reliably simulate and thus predict the physical world could become an additional consideration for the safety and alignment of strongly-capable AI agents, since they could in principle take more informed actions on the basis of a simulated understanding of the world.

\section{Methods}

\subsection{Physics task complexity}

As a preliminary step to designing research-level physics problems to evaluate LLMs, we find it useful to categorize physics problems into distinct complexity classes:

\begin{itemize}
    \item Class I - The problem calls for a non-parametric, unique and definite answer (e.g., value of a specific quantity of interest, time to an event, etc..)
	\item Class II - The problem calls for a parametric answer. That is, a satisfactory answer must consider the dependence on one or more explicit parameters of the system (e.g., the global stability of planetary system depends on the masses and detailed orbital configurations of individual planets, plus the time window of interest). This class includes the possibility that a parameter space survey is required to provide a quantitative answer.
	\item  Class III -   In this case, for a valid solution to be obtained, additional (implicit) physics beyond what is explicit in the problem statement must be considered. This touches on physics being a reductionist science in which order-of-magnitude arguments are often used to narrow down on a minimalistic model system that sufficiently captures the phenomenology of interest (e.g., computing satellite orbits around Earth could safely ignore or may be required to account for the Moon's gravity, depending on details on the satellite problem being considered). This is a challenging, arguably research-level capability that requires a strong physics world model, including an understanding of the modeling hierarchy in physics (e.g., in what regimes does one require classical vs. quantum vs. relativistic descriptions).
 	\item Class IV -  Out-of-distribution problems	- In this case, the problem is rarely discussed in the scientific literature and/or normally does not occur in nature. Nevertheless, it can be addressed using physical laws and principles. This exemplifies a strong form of generalization in physics \citep[e.g, hypothetical variations in fundamental constants can be used to reason about the Anthropic principle;][]{1979Natur.278..605C,Adams_2016}.
\end{itemize}

Most existing evaluation problems in the AI literature are class I because they admit a definite answer based on a well-defined ground truth, from which quantitative performance metrics can be computed \citep[e.g.,][]{hendrycks2021measuring,rein2023gpqa,mialon2023gaia}. By contrast, most research-level problems do not have unique, definite answers and it is frequent for several acceptable answers to receive consideration (until further scrutiny and scientific debate converge to a consensus view). At the graduate (PhD) education level, one may expect students to perform satisfactorily on class I-III and possibly class IV problems. 

We design our problems using the above categorization as general guidance, and attempt to cover the various levels of task complexity described by classes I-IV above.

\subsection{Code base essentials}
All four code bases were chosen on the basis of their open source nature, extensive online documentation, wide adoption by the research community, and the familiarity of the authors with them. 
\subsubsection{REBOUND}
\texttt{REBOUND} \citep{2012A&A...537A.128R,2020MNRAS.491.2885T} is, as stated in its documentation, ``an N-body integrator, i.e. a software package that can integrate the motion of particles under the influence of gravity. The particles can represent stars, planets, moons, ring or dust particles. \texttt{REBOUND} is very flexible and can be customized to accurately and efficiently solve many problems in astrophysics.''
\texttt{REBOUND} is written in C, with both C and Python APIs. Below is a minimum working example that illustrates its basic functionality:

\begin{mintedbox}[]{python}
import rebound
sim = rebound.Simulation()
sim.add("Sun")
sim.add("Jupiter")
sim.add("Saturn")
sim.integrate(100)
for orbit in sim.orbits():
    print(orbit)
\end{mintedbox}
In this code, we initialized a new rebound simulation in line 2, added the Sun, Jupiter, and Saturn on their current orbits in lines 3-5, then requested \texttt{REBOUND} to integrate (evolve) their orbits for 100 years in line 6, and print the final orbits in line 7-8.

\subsubsection{MESA}
\texttt{MESA} \citep{2011ApJS..192....3P} is a code that builds a model for the interior of a star, and then evolves it over time by solving the 1D fully-coupled structure and composition equations governing stellar evolution. It is written in Fortran although the user can in many cases setup problems using \texttt{Inlists} (\texttt{MESA}'s frontend) without requiring any coding at all. Fortran code can be written and invoked for additional physics or complex custom setups. 

Below is a minimum working \texttt{Inlist} example: 

\begin{mintedbox}[]{python}
&star_job

  ! begin with a pre-main sequence model
    create_pre_main_sequence_model = .true.

&controls

  ! starting specifications
    initial_mass = 15 ! in Msun units
    initial_z = 0.02

    ! stop when the star nears ZAMS (Lnuc/L > 0.99)
    stop_near_zams = .true.
\end{mintedbox}
Here we instructed \texttt{MESA} to evolve the early phases of a 15 solar masses (line 9), solar metallicity (line 10) star,  from pre-main sequence (line 4) till it reaches zero-age main sequence (line 13).

\subsubsection{Dedalus}
\texttt{Dedalus} is flexible open-source framework for solving arbitrary partial differential equations in arbitratry dimensions, using spectral methods. Dedalus uses a Python API and highly efficient pre-compiled scientific libraries for the back-end. Key features are an easy-to-use interface with symbolic equation entry and automated domain-decomposition for hands-free MPI-based parallelization in multi-dimensions.
Below is a minimal working \texttt{Dedalus} implementation of a straightforward 1D diffusion problem:

\begin{mintedbox}[]{python}
import numpy as np
from dedalus import public as de 

# Create the domain:
# use Chebyshev spectral decomposition on finite domain x=[0,1]
# also set numerical precision
x_basis = de.Chebyshev('x', 256, interval=(0, 1)) 
domain = de.Domain([x_basis], np.float64)

# Define the problem:
# initial value problem with two fields, u and ux
problem = de.IVP(domain, variables=['u', 'ux']) 

# Problem equations:
# diffusion equation 
problem.add_equation("dt(u) - dx(ux) = 0")
problem.add_equation("ux - dx(u) = 0") 

# Problem boundary conditions:
# zero flux on each domain side
problem.add_bc("left(ux) = 0") 
problem.add_bc("right(ux) = 0")

# Set the initial condition 
x = domain.grid(0)
u = solver.state['u']
u['g'] = (x-0.5)**2 

# Set up the solver
# use advanced Runge-Kutta time-stepper
solver = problem.build_solver(de.timesteppers.RK443) 

# Stopping criterion
solver.stop_iteration = 1000

# Main loop: step forward in time on timestep dt 
while solver.proceed:
    solver.step(dt=1e-3)
    
\end{mintedbox}

The finite numerical domain is setup on lines 6-7, the problem is setup on lines 11, 15-16 and 20-21. The initial conditions are setup on line 27. The problem solver is built on line 31 and time-stepping proceeds on lines 37-38.

\subsubsection{Scipy}
\texttt{Scipy} ``provides algorithms for optimization, integration, interpolation, eigenvalue problems, algebraic equations, differential equations, statistics,...''. Here we use it exclusively as an Ordinary Differential Equation (ODE) solver.  It is built on top of Fortran and C numerical computation libraries. \texttt{GPT4} uses the \texttt{scipy.integrate.odeint}\footnote{https://docs.scipy.org/doc/scipy/reference/generated/scipy.integrate.odeint.html\#scipy.integrate.odeint} method in our various evaluation problems.

\subsection{Problem Design}

We create $\sim 50$ original and challenging problems. As a baseline for these problems, we consider a hypothetical scenario in which PhD students have just taken a graduate-level course on a specific physics topics, including a computational component that introduced them to the standard numerical tools in use by the research community of interest. The problems we design could plausibly be elements of a final take-home exam for such a course.\footnote{Collectively, the authors have taught or practiced computational research in the four sub-domains covered by our problems.}

Some of the problems we design are considerably more challenging than others, as might be expected on a final take-home exam. Distinctly from a regular graduate exam, however, in the present work we need to pay close attention to the possibility that an LLM has seen and memorized numerous relevant code solutions as part of its extensive code pre-training. To minimize the risk of such data contamination, we avoid standard problems and instead contribute original problems crafted specifically for this work. In other words, our problems are designed to elicit some level of physics generalization capabilities from the LLMs. 
We note that our problem generation process is strongly shaped and biased by the expertise and research/teaching experience of both authors. Our limited set of 47 problems is therefore more illustrative than representative of potential computational physics problems that one could come up with in this context. A more diverse and uniform set of problems could presumably be obtained by enlisting a larger group of domain-experts to follow a careful problem generation methodology.

Nonetheless, we make an attempt to cover a broad range of physics situations within each sub-domain covered, adopting the following domain-specific principles for problem design:
\begin{itemize}
    \item Stellar Physics with \texttt{MESA}: we design problems at a research-level beyond that encountered in graduate-level textbooks. We focus on topics such as stellar lifetime, energy transport, internal composition, and surface winds. The examples cover the stellar life-cycle through the post main-sequence phase. The problems are designed such that while the simpler ones can be solved with \texttt{inlist}-flags and data post-processing alone, while the more complex ones necessitate the manual implementation of extra-physics routines and/or stopping conditions (in Fortran), which is commonly done for research with \texttt{MESA}. Out-of-distribution examples cover the case of stars with exotic compositions.    
    \item Celestial Mechanics with \texttt{REBOUND}: we focus on recurrent (exo)planetary dynamical problems often encountered in modern research settings and the academic literature, such as planetary migration, dynamical stability, and mean motion resonances. We design research-grade examples covering both the solar system and exoplanetary systems, and in some cases mix both as a test. No information about the initial orbits of the planets is ever provided unless absolutely necessary, allowing us to further test the basic information-recall capacity of LLMs in this setting. Out-of-distribution examples are also included, such as solvable cases with distraction features (electrically charged planets). 
    \item  1D Fluid Dynamics with \texttt{Dedalus}: many basic fluid phenomena are exhibited in 1D and can be conveniently decomposed into advection, diffusion (parabolic PDE) and wave (hyperbolic PDE) processes. We focus on these three idealized scenarios separately, even though more realistic scenarios often involve all three acting in concert in a given fluid system. We are also  guided by designing non-trivial problems that require generalization beyond standard code base and textbook examples (e.g. sink and source terms, shock-wave identification, non-linearity). Out-of-distribution cases include arbitrary time-dependent wave velocity and anti-diffusion.
    \item Non-linear dynamics with \texttt{SciPy}: we focus on the Lorenz dynamical system to leverage the extensive knowledge and code base for this well-understood non-linear dynamical system \citep{strogatz:2000}. We design non-trivial questions that require an understanding of the specific Lorenz system and of general principles in deterministic chaos, with a need to generalize beyond the textbook-level system.
\end{itemize}
    
\subsection{Prompt design}
\label{sec:prompt}

Rather than optimizing our prompts for improved conditional generation, we opt for a simple prompt formulation that approximates how the problem would be formulated in an academic setting. Our prompt is meant to sufficiently specify the physics
problem at hand for a graduate student with appropriate textbook and code knowledge to complete the assignment. We also add a few LLM specific elements to our prompts, by requesting a full code and specifying the software package and version that should be used to solve the problem (this prompt element is kept the same for all problems in a given problem class). Details on our inference pipeline are discussed in \S~\ref{sec:pipeline}{, and a detailed review of LLM prompting techniques can be found in \cite{promptreview}.}

\subsection{Solution Evaluation}

It is a priori unclear which method is best suited to evaluate the code solutions generated by LLMs in our study. The standard unit test methods often used in code
evaluation \citep[e.g.,][]{chen2021evaluating} are not appropriate here, since there is a continuum of possible code outputs that would qualify as successful and
satisfactory solutions. Even the propensity for the code generated to run without error is not an adequate measure since a single (minor) code error would not necessarily strongly devalue an otherwise involved and correct computational physics solution. 

Lacking hard metrics, we opt for several soft metrics that remain informative as to the quality of the solutions generated. First, we focus our evaluations on code lines, largely excluding any extra text and code comments generated.\footnote{Code comments are likely to be important in a more holistic evaluation framework.}

Our code evaluation relies on four metrics applied to each line of code generated:
\begin{itemize}
    \item Is the code correct (code syntax, logic and semantic)?
    \item Is the physics reasoning behind the code solution correct?
    \item Is the code line necessary as a solution element to the problem?
    \item Is the code line sufficient as a solution element to the problem?
\end{itemize}

These four criteria form the basis of our quantitative count of errors, under four separate categories (C: coding, P: physics, U: unnecessary, I: insufficient).

We recognize that the identification and classification of errors along those lines is subject to interpretation by the evaluator. We have made an effort to provide consistency in our evaluations. It is likely that some errors were missed despite best efforts by the evaluators. Our error counts could thus be interpreted as informed lower limits rather than rigorous numbers. It is our view that the statistics for error counts provides a reasonable global view of the frequency and type of errors generated in the solutions we evaluated.

A second approach to our evaluation method is to provide an overall, academic-like, grade for the solution. The motivation for this alternate grade is that code and (especially) physics errors come in various magnitudes. Some of the
physics errors we have reviewed, for instance, imply a serious lack of understanding of the problem as posed. As such, we assign one of three possible grades to all the solutions evaluated: Fail, Pass - and Pass +. Those grades are to be understood as academic-like evaluations of potential graduate student solutions for these problems under reasonable examination conditions (say, 30 min per problem, in an open-book setting).

We have not systematically verified that solutions run or generate reasonable outputs, as this is not part of our evaluation strategy. Indeed, we expect most solutions to not produce any outputs, or outputs that are not adequate for the problem at hand (as per our error analysis). Illustrative examples of code, physics, unnecessary and insufficient errors are presented via GPT4 solution excerpts in \S~{sec:showcase}.

No detailed attempt was made to calibrate errors across evaluators: author MAD designed and evaluated the \texttt{REBOUND} and \texttt{MESA} problems; author KM designed and evaluated the \texttt{Dedalus} and \texttt{Scipy} problems. Informally, both authors reviewed all the problems generated, as well as a subset of the evaluations
performed by the other author. We recognize that our problem design and evaluation
method falls short of the standards in the ML literature attempting to guarantee diverse and unbiased dataset and evaluation metrics. This shortcoming in
our work is largely due to the inherent challenge in generating original, meaningful problems in a highly specialized academic domain and the significant time-commitment required by the evaluation stage of our work.

\subsection{Pipeline}
\label{sec:pipeline}

We use the OpenAI python package to run our inferences, with \texttt{gpt-4-0314} (cutoff date of September 2021) as the LLM. We set the ``\texttt{temperature}'' parameter to 0, and leave the other parameters to their default values. The inferences were all run on Oct 25th 2023. Note that other versions of \texttt{GPT4} were briefly tested as well, including the new ``turbo'' version (\texttt{gpt-4-1106-preview}), but no obvious improvements were found  for our purposes compared to the version used for our work.  The pipeline generates a final composite prompt and feeds it to the API using two ``roles'' (system and user)  and the following template:
\begin{itemize}
    \item \texttt{system}: You are a useful coding assistant for a scientist.
    \item \texttt{user}: You help solve specific physics problem by writing python  code that makes use of domain-specialized python packages, as requested by the scientist. Your code is thoroughly commented for clarity and to provide descriptions of the various physics and coding choices made in your solution. You should always follow precisely the given instructions without omissions. \\
Prompt:$<$...$>$
\end{itemize}
The specific prompt instructions template for each code base are as follows:
\begin{itemize}
    \item {\texttt{MESA}: } Solve the following problem by providing the complete inputs to the latest version of the MESA stellar physics code (\texttt{inlist}\_project file setting the problem's options and parameters, and run\_star\_extras.f fortran file adding extra physics). You should also provide the python data analysis and postprocessing code needed to generate a complete final solution from MESA's .data output files.\\
Problem:$<$...$>$
    \item {\texttt{REBOUND}: } Use the REBOUND N-body integration Python package version 3.12.2, and its extension REBOUNDx version 3.1.0 to solve the following problem:  $<$...$>$
    \item {\texttt{Dedalus}: } Problem:$<$...$>$. Use the Dedalus PDE solver Python package, version 2. Give me a fully working code. 
    \item {\texttt{SciPy}: } Problem:$<$...$>$. Use the ODE integrator from the \texttt{SciPy} python package. Give me a fully working code.     
\end{itemize}

\section{Results}

\subsection{Statistics and Trends}

None of the code segments generated by \texttt{GPT4} are fully satisfactory solutions to the problems we posed. Nevertheless, the solutions have partial value, that we attempt to quantify with our soft metrics.

\begin{figure}[h]
    \centering
    \includegraphics[width=0.75\textwidth]{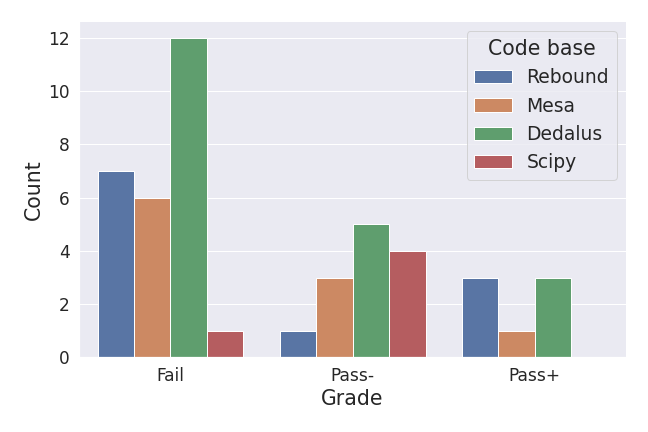}
    \caption{Histograms of \texttt{GPT4} solution grades, with either Fail, Pass- and Pass+ grades, grouped by code base. About $40\%$ of the solutions receive a passing grade, with significant variations across code bases. This soft metric focuses on the solution addressing the key physics ingredients for each problem. All solutions contain physics and/or coding errors.}
    \label{fig:one}
\end{figure}

We find that about $70-90 \%$ of all the code lines produced by \texttt{GPT4} are in fact necessary, sufficient and correct (coding \& physics). Typically, the solutions contain 25-50 lines of codes and exhibit several physics (P) and a few coding (C) errors, with some unnecessary or insufficient code lines interspersed.
Table~1 shows the summary statistics of errors and pass grades, per problem class. Fig~\ref{fig:one} shows an histogram of grades, grouped by code base.

We observe significant variations across problem classes and difficulty. The strongest performance is observed on non-linear dynamics problems with {\tt SciPy}\textcolor{black}{, but this problem class has significantly less examples than the others. The performance on problems in stellar physics with {\tt MESA}, fluid dynamics with {\tt Dedalus}, and celestial mechanics with {\tt REBOUND} are broadly comparable, despite fairly distinct code bases and background physics.} Higher complexity class does seem to correlate with weaker performance and possibly more frequent generation of placeholder code, although we do not have reliable statistics to make any quantitative statement. The relatively small numbers of U (and to some extent I) errors indicates that code generation is generally concise and efficient.

On the softer pass/fail metric, we estimate that about 40\% of the solutions generated could plausibly get a passing grade, in the sense that they contain enough essential physics and coding elements to marginally pass a take-home exam (see Fig~\ref{fig:one} for details). 

{We additionally performed an inter-rater test to check the level of agreement between the 2 authors. Each chose 4 examples of their corresponding code bases for the other author to review. We found a very high degree of agreement on the categorized errors of over 90\%. Only minor differences were noted, mainly some Unnecessary lines of code that were missed, in addition to a few that were deemed insufficient by one author but not the other. {We note that, while there are more sophisticated ways of checking the inter-rater agreement such as Cohen's kappa, we find our simple approach sufficient for this work.}}

Perhaps the most interesting result to come out of our evaluation work is that we are able to identify seemingly consistent failure modes of \texttt{GPT4} in computational physics. The most prominent flaws, some identified also during exploration (Appendix \ref{app:explore}), include:

\begin{itemize}
    \item a poor performance in dealing with physical units. This includes errors in simple conversions from one unit system to another, but also some apparent confusion between code units and physical units.
    \item a poor handling of code versioning. \texttt{GPT4} may arbitrarily pick a version that is not necessarily the latest available at its training cutoff date and can inconsistently invoke features that are not available for all relevant code versions.
    \item a tendency to {hallucinate\footnote{Hallucination refers to the generation of coherent but factually incorrect or nonsensical text by LLMs, which can occur when the model combines learned patterns in unexpected ways \citep{llmreview1,llmreview2}. } } plausibly-named sub-modules and functions. In an apparent effort to satisfy direct prompt requests or effective constraints on problem solutions, \texttt{GPT4} can hallucinate submodules or functions that seem appropriately named, but simply do not exist in the code base. \texttt{GPT4} can also hallucinate plausible-looking but ultimately false physical formulas.
    \item a lack of physical justification for global run parameters (e.g., simulation time, time-stepping choice, or upper-lower bounds for parametric exploration).
    \item an inability to define steady-state or stopping conditions reliably for time-dependent simulations.
    \item a tendency to use common approximate equations beyond their physical applicability range.
\end{itemize}

Besides these trends emerging across various problems, we also collect below more detailed observations about some of the most recognizable (physics and coding) errors made by GPT4, grouped by problem class.


\begin{table}[] 
\resizebox{\columnwidth}{!}{%
\begin{tabular}{rccccc} 
\cline{2-6}
\multicolumn{1}{l|}{} &
  \multicolumn{1}{l|}{{C: code error}} &
  \multicolumn{1}{l|}{{P: physics error}} &
  \multicolumn{1}{l|}{{I: insufficient error}} &
  \multicolumn{1}{l|}{{U: unnecessary error}} &
  \multicolumn{1}{c|}{{Pass Ratio}} \\ \hline
\multicolumn{1}{|r|}{{\tt REBOUND}} &
  \multicolumn{1}{c|}{1.86±0.83} &
  \multicolumn{1}{c|}{1.50±0.71} &
  \multicolumn{1}{c|}{1.57±0.49} &
  \multicolumn{1}{c|}{2.25±1.64} &
  \multicolumn{1}{c|}{0.36} \\ \hline
\multicolumn{1}{|r|}{{\tt Dedalus}} &
  \multicolumn{1}{c|}{1.75±0.92} &
  \multicolumn{1}{c|}{2.59±1.33} &
  \multicolumn{1}{c|}{1.20±0.40} &
  \multicolumn{1}{c|}{1.83±1.07} &
  \multicolumn{1}{c|}{0.40} \\ \hline
\multicolumn{1}{|r|}{{\tt MESA}} &
  \multicolumn{1}{c|}{2.50±2.01} &
  \multicolumn{1}{c|}{1.50±0.50} &
  \multicolumn{1}{c|}{1.00±0.00} &
  \multicolumn{1}{c|}{0±0.00} &
  \multicolumn{1}{c|}{0.40} \\ \hline
\multicolumn{1}{|r|}{{\tt SciPy}} &
  \multicolumn{1}{c|}{1.00±0.00} &
  \multicolumn{1}{c|}{1.20±0.40} &
  \multicolumn{1}{c|}{1.40±0.49} &
  \multicolumn{1}{c|}{1.00±0.00} &
  \multicolumn{1}{c|}{0.80} \\ \hline
\multicolumn{1}{|r|}{{Overall Mean}} &
  \multicolumn{1}{c|}{{1.78}} &
  \multicolumn{1}{c|}{{1.70}} &
  \multicolumn{1}{c|}{{1.29}} &
  \multicolumn{1}{c|}{{1.27}} &
  \multicolumn{1}{c|}{{0.49}} \\ \hline
\multicolumn{1}{l}{} &
  \multicolumn{1}{l}{} &
  \multicolumn{1}{l}{} &
  \multicolumn{1}{l}{} &
  \multicolumn{1}{l}{} &
  \multicolumn{1}{l}{} \\ \hline
\multicolumn{1}{|r|}{{\tt REBOUND}} &
  \multicolumn{1}{c|}{2} &
  \multicolumn{1}{c|}{1} &
  \multicolumn{1}{c|}{2} &
  \multicolumn{1}{c|}{1.5} &
  \multicolumn{1}{c|}{0.36} \\ \hline
\multicolumn{1}{|r|}{{\tt Dedalus}} &
  \multicolumn{1}{c|}{1.50} &
  \multicolumn{1}{c|}{3.00} &
  \multicolumn{1}{c|}{1.00} &
  \multicolumn{1}{c|}{1.00} &
  \multicolumn{1}{c|}{0.40} \\ \hline
\multicolumn{1}{|r|}{{\tt MESA}} &
  \multicolumn{1}{c|}{2} &
  \multicolumn{1}{c|}{1.5} &
  \multicolumn{1}{c|}{1} &
  \multicolumn{1}{c|}{0.00} &
  \multicolumn{1}{c|}{0.40} \\ \hline
\multicolumn{1}{|r|}{{\tt SciPy}} &
  \multicolumn{1}{c|}{1.00} &
  \multicolumn{1}{c|}{1.00} &
  \multicolumn{1}{c|}{1.00} &
  \multicolumn{1}{c|}{1.00} &
  \multicolumn{1}{c|}{0.80} \\ \hline
\multicolumn{1}{|r|}{{Overall Median }} &
  \multicolumn{1}{c|}{{1.75}} &
  \multicolumn{1}{c|}{{1.25}} &
  \multicolumn{1}{c|}{{1.00}} &
  \multicolumn{1}{c|}{{1.00}} &
  \multicolumn{1}{c|}{{0.40}} \\ \hline
\end{tabular}%
\label{tab:one}
}
\caption{Summary statistics of error types across our various problems (grouped by code base). The upper table shows mean + standard deviation computed over all problems within a given code base. The lower table shows the median statistics. The fraction of solutions with a passing grade is shown in the last column (see also Fig~1 for grade histograms.)}
\end{table}

\subsection{Error showcasing}
\label {sec:showcase}
\subsubsection{\texttt{Dedalus}}

The following code excerpt is from a \texttt{Dedalus} advection problem. Code line 5 is labeled as insufficient (I) in failing to define or compute tolerance on physical grounds.
\begin{mintedbox}[]{python}
# Parameters
v = 0.1 # Advection velocity
L = 1 # Spatial domain length
FWHM = 0.3 # Full-Width-at-Half-Maximum 
S = 0.01 # Peak rate amplitude
tolerance = 1e-6 # Steady state tolerance
\end{mintedbox}

The following code excerpt is from a \texttt{Dedalus} advection problem. Code lines 2 and 3 are both labeled as physics (P) errors in failing to set zero-flux ($dx(u)=0$) boundary conditions (referred to as outflow boundary conditions in the problem statement).
\begin{mintedbox}[]{python}
# Set the outflow boundary conditions
problem.add_bc("right(u) = 1") 
problem.add_bc("left(u) = 1")
\end{mintedbox}

The following code excerpt is from a \texttt{Dedalus} diffusion problem. Code line 4 is labeled as a coding (C) error. \texttt{GPT4} hallucinates a plausibly-named 'set\_initial' function that does not exist in the \texttt{Dedalus} code base.
\begin{mintedbox}[]{python}
# Set the initial condition
x = domain.grid(0)
u0 = np.exp(-4 * np.log(2) * (x - 0.5)**2 / 0.3**2) 
problem.set_initial('u', u0)
\end{mintedbox}

The following code excerpt is from a \texttt{Dedalus} acoustic wave problem. Code line 2 is labeled as unnecessary (U) as the defined parameter $a$ is never used anywhere else in the generated code. Code line 3 is also labeled as a physics (P) error since \texttt{GPT4} sets up a diffusion (parabolic) equation rather than wave (hyperbolic) equation. 
\begin{mintedbox}[]{python}
problem = de.IVP(domain, variables=['u', 'ux']) 
problem.parameters['a'] = a 
problem.add_equation("dt(u) - dx(ux) = 0") 
problem.add_equation("ux - dx(u) = 0")
\end{mintedbox}

The following code excerpt is from a \texttt{Dedalus} diffusion problem. One of the three code lines 3-5 is labeled as insufficient (I) because \texttt{GPT4} fails to define a finite stopping time for the simulation. Assigning a finite value to any one of these lines would avoid running a never-ending simulation.   
\begin{mintedbox}[]{python}
# Set up the solver
solver = problem.build_solver(de.timesteppers.RK443) 
solver.stop_sim_time = np.inf
solver.stop_wall_time = np.inf
solver.stop_iteration = np.inf
\end{mintedbox}

\subsubsection{\texttt{SciPy}}

The following code excerpt is from a \texttt{SciPy} non-linear dynamics problem. Code line 4 in the 'objective' function is labeled as a physics (P) error as \texttt{GPT4} fails to setup an initial condition on the z-axis (the variable assignment should be [0,0,25]). Code line 5 in the  'objective' function is labeled as insufficient (I) as \texttt{GPT4} selects a time span that is arbitrary (unjustified) and a priori not based on any physical or dynamical argument. 
\begin{mintedbox}[]{python}
# Define the objective function to minimize the time it takes for the system to cross the y-z plane twice
def objective(params):
    A, omega = params
    initial_state = [1, 1, 25] # Specific initial condition on the z axis with z=25 
    t_span = (0, 100) # Time span for integration
\end{mintedbox}

\subsubsection{\texttt{REBOUND} }

The following code excerpt is from a \texttt{REBOUND} problem and illustrates a recurrent issue with \texttt{GPT4} generations to \texttt{REBOUND} problems. Here \texttt{GPT4} uses IAS15 as the integrator (code line 1), but then in code line 2 defines a timestep even though IAS15 automatically defines and adapts its timestep. Code line 2 is labeled as unnecessary (U). Moreover, note that even if one had to specify a timestep, the value proposed by the LLM (0.5) is too high.  

\begin{mintedbox}[]{python}
sim.integrator = "ias15"
sim.dt = 0.5 * sim.particles[1].P  # half of Neptune's period
\end{mintedbox}

\subsubsection{\texttt{MESA}}

The following code excerpt is from a (\texttt{MESA}) problem. Code line 8 is labeled as a physics (P) error as GPT4 defines a non-existent (and physically non-sensical) stellar mass loss rate equation.    

\begin{mintedbox}[]{python}
  ! Calculate the wind mass loss rate using an appropriate formula
  ! For example, you can use the Reimers formula or Vink et al. (2001) formula
  ! Here, we use a simple example formula: wind_mass_loss_rate = C * mass^2 / radius
  ! where C is a constant that you can adjust to match the desired wind mass loss rate

  real(dp) :: C
  C = 1d-13
  wind_mass_loss_rate = C * mass**2 / radius 
\end{mintedbox}

The following code excerpt is from a (\texttt{MESA}) problem. Code line 14 is labeled as I (insufficient) as \texttt{GPT4} defines a non-existent stopping condition flag in the \texttt{inlist}, instead of implementing it manually. In addition, code lines 7, 9 and 11 are labeled as C (coding) erros since \texttt{GPT4} uses non-existing control flags (that do exist, but under different names). 

\begin{mintedbox}[]{python}
&controls
  ! Set the initial metallicity
  initial_z = 0.02
  ! Set the mixing length theory parameter
  mixing_length_alpha = 1.5
  ! Set the maximum number of steps
  max_num_steps = 10000 
  ! Set the maximum number of retries
  max_num_retries = 100  
  ! Set the maximum number of backups
  max_num_backups = 100  
  
  ! Set the stopping condition for carbon burning
  stop_at_carbon_burning = .true.
\end{mintedbox}

\subsection{Additional observations: \tt REBOUND}
\begin{itemize}
    \item For all parametric problems where the LLM was asked about an object x ``placed between planet a and planet b'', \texttt{GPT4} used the arithmetic mean of the semimajor axis of planets a and b as the location of object x. At no point was this quantity treated as an actual explorable parameter as it should be.  
    \item For all orbital stability questions, \texttt{GPT4} always checked for instabilities by looking for any close encounter in the system. This is problematic since (i) the flagged close encounter might not be relevant to the specific planet/object in question, and (ii) a close encounter does not necessary imply orbital instability.
    \item For all mean motion resonance (MMR) related problems, \texttt{GPT4} always searched for resonances by checking whether the periods are a simple integer ratio of each other, but never for formal conditions such as librating resonant angles. This can be misleading as two objects can have integer-ratio periods without actually being trapped in a resonance. 
    \item Contrary to Solar System planets, \texttt{GPT4} failed to get the correct orbital parameters of exoplanets (mass, semimajor axis, eccentricity, etc.) when provided with the name of the planet only. (Circumstantial evidence suggests that this can be remedied by referencing specific popular databases). 
   
\end{itemize}

\subsection{Additional observations: \tt MESA}
\begin{itemize}
    \item Current generation LLMs are not capable of generating \texttt{MESA} \texttt{\texttt{inlist}} with correct syntax, even though they seem familiar with the code and its uses and principles. \texttt{inlist}s are technically not code, but rather a list of flags and parameters. While \texttt{GPT4} seems to deal with this, it often used non-existing flags with reasonably chosen names, rather than the correct ones, or placed the flags under the wrong \texttt{inlist} sub-category. This usually results in \texttt{GPT4} reducing the solution of a complex problem to a non-existent flag instead of actually solving the problem. 
    \item  \texttt{GPT4} can however distinguish most of time between what goes in the \texttt{inlist} as a flag and what should go into \texttt{run\_star\_extras.f} files as Fortran code. 
    \item  When specific termination conditions were necessary, \texttt{GPT4} often just used non-existing \texttt{inlist} termination flags, reasonably named for the task. The termination criteria sometimes did exist but under a different flag name, although in many cases the termination conditions should have been manually implemented.
    \item When \texttt{MESA} output post-processing was necessary to get a final answer, the corresponding python code generated by \texttt{GPT4} was usually adequate. This highlights further the rift between GPT4's capability in coding vs setting the \texttt{inlist}'s flags. 
    \item \texttt{GPT4} automatically invoked mixing length theory flags that were necessary when solving convection-related problems.
    \item In problems with no explicit stopping criteria, the ones chosen by \texttt{GPT4} were often arbitrary. 
    \item \textcolor{black}{Similarly to the \texttt{REBOUND} cases, \texttt{GPT4} had a tendency with \texttt{MESA} to fix any implicit parameter to a constant value rather than exploring a relevant range of values. }
    
\end{itemize}

\subsection{Additional observations: \texttt{Dedalus}}

\begin{itemize}
    \item \texttt{GPT4} is frequently able to load appropriate modules and broadly set up a problem (i.e. define the domain, variables and parameters, time-stepping, etc...) under the \texttt{Dedalus} code base, so that errors tend to be narrower and more specific than the general problem setup.

    \item GPT4 is prone to code-base specific hallucinations, by invoking sub-modules and functions (or related attributes) that are non-existent. These hallucinated features are plausibly named, given the context in which they are invoked, which means careful code documentation examination maybe be needed to rule out their existence. 

    \item With limited available statistics, it appears that GPT4 performs better at diffusion than advection problems, and worse at wave problems.  

    \item Some of the physics errors seem to have a stochastic component, in the sense that an error may be present in one problem but absent from another related problem, even though both problems call for the same line of code. In addition, some of the most blatant physics errors would not be expected at a graduate level of performance.

    \item GPT4 fails or struggles to define physically-justifiable run-time for time-dependent problems. Ideally, this would be computed from global run parameters and other general considerations. GPT 4 also fails to define or identify good measures of steady-state in time-dependent problems.

    \item  GPT4 fails to justify on physical grounds the values of global run parameters, such as sensible bounds and steps in a systematic parameter-sspace exploration.
    
\end{itemize}

\subsection{Additional observations: \tt SciPy}
\begin{itemize}
    \item GPT4 performs relatively well on problems related to the Lorenz system of equations. With limited available statistics, it seems to be the best performance among the four classes of problems considered.

    \item GPT4 performs strongly at defining algorithmic iterations and optimizations (e.g. to find an optimum over runs with varying parameters).

    \item With limited available statistics, it seems that errors on problmes related to the Lorenz system are more deterministic than the stochastic errors above-mentioned for the \texttt{Dedalus} code base. 
\end{itemize}

\subsection{Context-specific problems}

Most of our problems are generic physics problems, without any context-specific information.  We have injected some general contextual information in a subset of problems, using planet names as specific contextual anchors.

Under the specific \texttt{REBOUND} code base,  we find that \texttt{GPT4} is generally able to use contextual information on Solar System objects (i.e., infer relevant physical attributes from the name only) but that this capability does not reliably extend to less-well documented extrasolar planets. \textcolor{black}{With \texttt{REBOUND}, we further test contextual inputs by inserting  additional solar system objects and giving them (potentially confusing) known exoplanetary names in some problems, but found this to have no obvious effect on the output. Under the specific \texttt{MESA} code base, we use random floating point numbers (26.1451) or mathematical constants (3.1416) for stellar masses, as contextual cues, which also had no obvious effect on the output. }

In addition, we note all our non-linear dynamics problems under the \texttt{Scipy} code base are variations around the standard Lorenz system. We find that \texttt{GPT4} is reliably able to code the standard Lorenz dynamical system of equations from scratch, using the system name as sole context. 

It may be interesting to more systematically explore the context-awareness of LLMs in physics and science.

\section{Limitations and Extensions}

Clearly, our work provides only a very preliminary and limited view of the general computational physics capabilities of LLMs, one that could be expanded and improved upon in a variety of ways.\footnote{For reference, we estimate that the problem and evaluation design phase of our analysis required $\sim 20$ hours of work and the solution evaluation phase another $\sim 10$ hours. \texttt{GPT4} generated all 47 code solutions in about $15$ minutes.} We highlight a few potential directions for future work here:

\begin{itemize}
    \item It would be beneficial to carry out a more extensive
and systematic study to garner more reliable statistics and understanding of the nature of LLM errors in the specific computational physics domains we have explored.

    \item  It would also be beneficial to explore further the physics challenge by introducing problems that go beyond the narrow ones covered here, e.g. by including multi-dimensionality, radiation transport and other common ingredients in research-level simulations .

    \item  It may be interesting to correlate the performance of LLMs with the extent to which open-source packages and code examples are available for pre-training. Previous work indicates that such correlations could be a significant factor in determining performance for any LLM generation task \citep{mccoy2023embers}.

    \item Attempts to systematically improve on the simple
prompt strategy we have adopted may also be of value (see Appendix~\ref{app:explore} for a relevant exploration).

    \item  It may be worthwhile to extend our benchmark to more conversational, human-in-the-loop settings or perhaps with the availability of extra tools for more agentic setups.

    \item  Extension beyond classical physics, to cover domains such as solid-state,  quantum physics, or relativistic physics, would also be of great interest. Presumably, similar general simulation efforts could be developed in computational biology, chemistry or engineering.

    \item  If training datasets, or reward functions for reinforcement learning, can be constructed on numerical simulation tasks, it would then be particularly interesting to fine-tune LLMs, in an attempt to further elicit their capabilities in computational science beyond the restrictive limits of prompting.

    \item  In-context learning may be another interesting avenue to pursue for improved performance on computational science tasks.

\end{itemize}

\section{Conclusion}

We presented an evaluation of LLMs on graduate-level to research-level computational physics problems. Our main results can be summarized as twofold.
First, \texttt{GPT4} is currently the best LLM for this task but it is not capable of autonomously generating code solutions at the graduate/research level we evaluate for. Second, \texttt{GPT4} exhibits consistent failure modes in the computational physics domain that suggest obvious targets for performance improvement and monitoring with future generation AI systems.

It is tempting to speculate on the origin of these failure modes. Some appear to be directly related to the LLM hallucination phenomenon (e.g., when invoking non-existent functions). {The poor ability to track package versions may be related to hallucinations and/or the limited time-awareness of LLMs given their time-agnostic next-token pre-training objective\footnote{LLMs are pre-trained on a massive “corpus” of text data. This data however is not structured chronologically. If the corpus contains two different versions of a code library, both are treated equally leading to LLMs that cannot easily distinguish between the two versions. }.} These two shortcomings could perhaps be (partially) addressed with general strategies to reduce hallucinations \citep{rawte2023survey,zhang2023sirens,huang2023survey} and to equip LLMs with an explicit sense of time regarding the text they are trained on \citep[e.g.,][]{llama2}. {Hallucinating non-existent sub-modules and functions could also possibly be mitigated by incorporating mechanisms to verify the existence of referenced modules and functions against known libraries or datasets during the problem-solving process.}

On the other hand, some of the failure modes we identified seem to be related to limitations of the physics world models learned by GPT4. {Unit conversion, for example, is a sub-graduate level challenge, even in the context of code-specific units, that could perhaps be addressed with a dedicated training corpus.} By contrast, challenges in defining steady-state, stopping criteria or global parameter values for simulations could be related to more fundamental limits in the reasoning and planning capabilities of GPT4 in the physics domain.

It will be interesting to determine to what extent future generation models improve upon GPT4's computational physics shortcomings, either as a simple byproduct of scaling or from adopting general training strategies aimed at reducing hallucinations and increasing reasoning and planning abilities. It could also be the case that tailored training strategies for computational physics are needed to overcome some of the shortcomings we have identified, since some of the common-sense and reasoning steps behind practical choices made in computational physics are not systematically documented in the academic literature (or in textbooks). 

\section*{Author Contributions}

MAD and KM conceived the project jointly and contributed equally to the design of physics problems and to the evaluation of LLM solutions. MAD designed and operated the LLM generation pipeline. KM designed the evaluation methodology, in consultation with MAD. Both authors contributed equally to the analysis of results and code examples. MAD prepared tables, figures and performed the exploration reported in Appendix~\ref{app:explore}. KM was lead-writer, with significant contributions from MAD.

\section*{Acknowledgements }
We thank the referees for their useful comments that helped improve this manuscript. 
MAD is supported by Tamkeen under the NYU Abu Dhabi Research Institute grant CASS. KM is supported by the National Science and Engineering Research Council of Canada.

\section*{Data availability statement}
All of the data and code used in this work are publicly available. The complete set of input (prompt) questions are provided within the manuscript. The code used to generate the LLM responses is available \href{https://github.com/malidib/llm-physics/blob/main/main_inference_llm.py}{here}.

\newpage

\bibliographystyle{mnras} 
\bibliography{stars.bib}

\begin{thebibliography}{}
\makeatletter
\relax
\def\mn@urlcharsother{\let\do\@makeother \do\$\do\&\do\#\do\^\do\_\do\%\do\~}
\def\mn@doi{\begingroup\mn@urlcharsother \@ifnextchar [ {\mn@doi@}
  {\mn@doi@[]}}
\def\mn@doi@[#1]#2{\def\@tempa{#1}\ifx\@tempa\@empty \href
  {http://dx.doi.org/#2} {doi:#2}\else \href {http://dx.doi.org/#2} {#1}\fi
  \endgroup}
\def\mn@eprint#1#2{\mn@eprint@#1:#2::\@nil}
\def\mn@eprint@arXiv#1{\href {http://arxiv.org/abs/#1} {{\tt arXiv:#1}}}
\def\mn@eprint@dblp#1{\href {http://dblp.uni-trier.de/rec/bibtex/#1.xml}
  {dblp:#1}}
\def\mn@eprint@#1:#2:#3:#4\@nil{\def\@tempa {#1}\def\@tempb {#2}\def\@tempc
  {#3}\ifx \@tempc \@empty \let \@tempc \@tempb \let \@tempb \@tempa \fi \ifx
  \@tempb \@empty \def\@tempb {arXiv}\fi \@ifundefined
  {mn@eprint@\@tempb}{\@tempb:\@tempc}{\expandafter \expandafter \csname
  mn@eprint@\@tempb\endcsname \expandafter{\@tempc}}}

\bibitem[\protect\citeauthoryear{{Abedi}, {Alshybani}, {Shahadat}  \&
  {Murillo}}{{Abedi} et~al.}{2023}]{2023arXiv230913059A}
{Abedi} M.,  {Alshybani} I.,  {Shahadat} M. R.~B.,   {Murillo} M.~S.,  2023,
  \mn@doi [arXiv e-prints] {10.48550/arXiv.2309.13059}, \href
  {https://ui.adsabs.harvard.edu/abs/2023arXiv230913059A} {p. arXiv:2309.13059}

\bibitem[\protect\citeauthoryear{Adams}{Adams}{2016}]{Adams_2016}
Adams F.~C.,  2016, \mn@doi [Journal of Cosmology and Astroparticle Physics]
  {10.1088/1475-7516/2016/02/042}, 2016, 042–042

\bibitem[\protect\citeauthoryear{Anthropic}{Anthropic}{2023}]{Anthropic_core}
Anthropic P.,  2023, Core Views on AI Safety: When, Why, What, and How, \url
  {https://www.anthropic.com/index/core-views-on-ai-safety}

\bibitem[\protect\citeauthoryear{{Bovy}}{{Bovy}}{2015}]{galpy1}
{Bovy} J.,  2015, \mn@doi [\apjs] {10.1088/0067-0049/216/2/29}, \href
  {https://ui.adsabs.harvard.edu/abs/2015ApJS..216...29B} {216, 29}

\bibitem[\protect\citeauthoryear{Bovy}{Bovy}{2024}]{galpy2}
Bovy J.,  2024, {Dynamics and Astrophysics of Galaxies, Princeton University
  Press, Princeton, NJ (in preparation)}

\bibitem[\protect\citeauthoryear{Bowman}{Bowman}{2023}]{bowman2023things}
Bowman S.~R.,  2023, Eight Things to Know about Large Language Models
  (\mn@eprint {arXiv} {2304.00612})

\bibitem[\protect\citeauthoryear{{Boyko} et~al.,}{{Boyko}
  et~al.}{2023}]{2023arXiv231104929B}
{Boyko} J.,  et~al., 2023, \mn@doi [arXiv e-prints]
  {10.48550/arXiv.2311.04929}, \href
  {https://ui.adsabs.harvard.edu/abs/2023arXiv231104929B} {p. arXiv:2311.04929}

\bibitem[\protect\citeauthoryear{Brown-Cohen, Irving  \& Piliouras}{Brown-Cohen
  et~al.}{2023}]{browncohen2023scalable}
Brown-Cohen J.,  Irving G.,   Piliouras G.,  2023, Scalable AI Safety via
  Doubly-Efficient Debate (\mn@eprint {arXiv} {2311.14125})

\bibitem[\protect\citeauthoryear{{Burns}, {Vasil}, {Oishi}, {Lecoanet}  \&
  {Brown}}{{Burns} et~al.}{2020}]{2020PhRvR...2b3068B}
{Burns} K.~J.,  {Vasil} G.~M.,  {Oishi} J.~S.,  {Lecoanet} D.,   {Brown} B.~P.,
   2020, \mn@doi [Physical Review Research] {10.1103/PhysRevResearch.2.023068},
  \href {https://ui.adsabs.harvard.edu/abs/2020PhRvR...2b3068B} {2, 023068}

\bibitem[\protect\citeauthoryear{{Carr} \& {Rees}}{{Carr} \&
  {Rees}}{1979}]{1979Natur.278..605C}
{Carr} B.~J.,  {Rees} M.~J.,  1979, \mn@doi [\nat] {10.1038/278605a0}, \href
  {https://ui.adsabs.harvard.edu/abs/1979Natur.278..605C} {278, 605}

\bibitem[\protect\citeauthoryear{Chen et~al.,}{Chen
  et~al.}{2021}]{chen2021evaluating}
Chen M.,  et~al., 2021, Evaluating Large Language Models Trained on Code
  (\mn@eprint {arXiv} {2107.03374})

\bibitem[\protect\citeauthoryear{Christiano, Xu  \& Cotra}{Christiano
  et~al.}{2021}]{ELK}
Christiano P.,  Xu M.,   Cotra A.,  2021, Eliciting latent knowledge: How to
  tell if your eyes deceive you., \url
  {https://www.alignmentforum.org/posts/qHCDysDnvhteW7kRd/arc-s-first-technical-report-eliciting-latent-knowledge}

\bibitem[\protect\citeauthoryear{{Fagbohun}, {Harrison}  \&
  {Dereventsov}}{{Fagbohun} et~al.}{2024}]{promptreview}
{Fagbohun} O.,  {Harrison} R.~M.,   {Dereventsov} A.,  2024, \mn@doi [arXiv
  e-prints] {10.48550/arXiv.2402.14837}, \href
  {https://ui.adsabs.harvard.edu/abs/2024arXiv240214837F} {p. arXiv:2402.14837}

\bibitem[\protect\citeauthoryear{Hendrycks, Burns, Basart, Zou, Mazeika, Song
  \& Steinhardt}{Hendrycks et~al.}{2021}]{hendrycks2021measuring}
Hendrycks D.,  Burns C.,  Basart S.,  Zou A.,  Mazeika M.,  Song D.,
  Steinhardt J.,  2021, Measuring Massive Multitask Language Understanding
  (\mn@eprint {arXiv} {2009.03300})

\bibitem[\protect\citeauthoryear{Huang et~al.,}{Huang
  et~al.}{2023}]{huang2023survey}
Huang L.,  et~al., 2023, A Survey on Hallucination in Large Language Models:
  Principles, Taxonomy, Challenges, and Open Questions (\mn@eprint {arXiv}
  {2311.05232})

\bibitem[\protect\citeauthoryear{Hubinger, Jermyn, Treutlein, Hudson  \&
  Woolverton}{Hubinger et~al.}{2023}]{hubinger2023conditioning}
Hubinger E.,  Jermyn A.,  Treutlein J.,  Hudson R.,   Woolverton K.,  2023,
  Conditioning Predictive Models: Risks and Strategies (\mn@eprint {arXiv}
  {2302.00805})

\bibitem[\protect\citeauthoryear{Kaddour, Harris, Mozes, Bradley, Raileanu  \&
  McHardy}{Kaddour et~al.}{2023a}]{llmreview2}
Kaddour J.,  Harris J.,  Mozes M.,  Bradley H.,  Raileanu R.,   McHardy R.,
  2023a, Challenges and Applications of Large Language Models (\mn@eprint
  {arXiv} {2307.10169})

\bibitem[\protect\citeauthoryear{Kaddour, Harris, Mozes, Bradley, Raileanu  \&
  McHardy}{Kaddour et~al.}{2023b}]{kaddour2023challenges}
Kaddour J.,  Harris J.,  Mozes M.,  Bradley H.,  Raileanu R.,   McHardy R.,
  2023b, Challenges and Applications of Large Language Models (\mn@eprint
  {arXiv} {2307.10169})

\bibitem[\protect\citeauthoryear{{Khan}, {Ramsdell}, {Falor}  \&
  {Karimi}}{{Khan} et~al.}{2023}]{2023arXiv231102640K}
{Khan} M. F.~A.,  {Ramsdell} M.,  {Falor} E.,   {Karimi} H.,  2023, \mn@doi
  [arXiv e-prints] {10.48550/arXiv.2311.02640}, \href
  {https://ui.adsabs.harvard.edu/abs/2023arXiv231102640K} {p. arXiv:2311.02640}

\bibitem[\protect\citeauthoryear{Liu et~al.,}{Liu
  et~al.}{2023}]{liu2023mlbench}
Liu Y.,  et~al., 2023, ML-Bench: Large Language Models Leverage Open-source
  Libraries for Machine Learning Tasks (\mn@eprint {arXiv} {2311.09835})

\bibitem[\protect\citeauthoryear{{Maslej} et~al.,}{{Maslej}
  et~al.}{2023}]{2023arXiv231003715M}
{Maslej} N.,  et~al., 2023, \mn@doi [arXiv e-prints]
  {10.48550/arXiv.2310.03715}, \href
  {https://ui.adsabs.harvard.edu/abs/2023arXiv231003715M} {p. arXiv:2310.03715}

\bibitem[\protect\citeauthoryear{McCoy, Yao, Friedman, Hardy  \&
  Griffiths}{McCoy et~al.}{2023}]{mccoy2023embers}
McCoy R.~T.,  Yao S.,  Friedman D.,  Hardy M.,   Griffiths T.~L.,  2023, Embers
  of Autoregression: Understanding Large Language Models Through the Problem
  They are Trained to Solve (\mn@eprint {arXiv} {2309.13638})

\bibitem[\protect\citeauthoryear{Mialon, Fourrier, Swift, Wolf, LeCun  \&
  Scialom}{Mialon et~al.}{2023}]{mialon2023gaia}
Mialon G.,  Fourrier C.,  Swift C.,  Wolf T.,  LeCun Y.,   Scialom T.,  2023,
  GAIA: a benchmark for General AI Assistants (\mn@eprint {arXiv} {2311.12983})

\bibitem[\protect\citeauthoryear{Michael, Mahdi, Rein, Petty, Dirani,
  Padmakumar  \& Bowman}{Michael et~al.}{2023}]{michael2023debate}
Michael J.,  Mahdi S.,  Rein D.,  Petty J.,  Dirani J.,  Padmakumar V.,
  Bowman S.~R.,  2023, Debate Helps Supervise Unreliable Experts (\mn@eprint
  {arXiv} {2311.08702})

\bibitem[\protect\citeauthoryear{Minaee, Mikolov, Nikzad, Chenaghlu, Socher,
  Amatriain  \& Gao}{Minaee et~al.}{2024}]{llmreview1}
Minaee S.,  Mikolov T.,  Nikzad N.,  Chenaghlu M.,  Socher R.,  Amatriain X.,
  Gao J.,  2024, Large Language Models: A Survey (\mn@eprint {arXiv}
  {2402.06196})

\bibitem[\protect\citeauthoryear{Morris, Sohl-dickstein, Fiedel, Warkentin,
  Dafoe, Faust, Farabet  \& Legg}{Morris et~al.}{2023}]{morris2023levels}
Morris M.~R.,  Sohl-dickstein J.,  Fiedel N.,  Warkentin T.,  Dafoe A.,  Faust
  A.,  Farabet C.,   Legg S.,  2023, Levels of AGI: Operationalizing Progress
  on the Path to AGI (\mn@eprint {arXiv} {2311.02462})

\bibitem[\protect\citeauthoryear{OpenAI}{OpenAI}{2023}]{GPT4V_card}
OpenAI 2023, GPT-4V(ision) System Card, \url
  {https://cdn.openai.com/papers/GPTV_System_Card.pdf}

\bibitem[\protect\citeauthoryear{{Paxton}, {Bildsten}, {Dotter}, {Herwig},
  {Lesaffre}  \& {Timmes}}{{Paxton} et~al.}{2011}]{2011ApJS..192....3P}
{Paxton} B.,  {Bildsten} L.,  {Dotter} A.,  {Herwig} F.,  {Lesaffre} P.,
  {Timmes} F.,  2011, \mn@doi [\apjs] {10.1088/0067-0049/192/1/3}, \href
  {https://ui.adsabs.harvard.edu/abs/2011ApJS..192....3P} {192, 3}

\bibitem[\protect\citeauthoryear{Rawte, Sheth  \& Das}{Rawte
  et~al.}{2023}]{rawte2023survey}
Rawte V.,  Sheth A.,   Das A.,  2023, A Survey of Hallucination in Large
  Foundation Models (\mn@eprint {arXiv} {2309.05922})

\bibitem[\protect\citeauthoryear{{Rein} \& {Liu}}{{Rein} \&
  {Liu}}{2012}]{2012A&A...537A.128R}
{Rein} H.,  {Liu} S.~F.,  2012, \mn@doi [\aap] {10.1051/0004-6361/201118085},
  \href {https://ui.adsabs.harvard.edu/abs/2012A&A...537A.128R} {537, A128}

\bibitem[\protect\citeauthoryear{Rein, Hou, Stickland, Petty, Pang, Dirani,
  Michael  \& Bowman}{Rein et~al.}{2023}]{rein2023gpqa}
Rein D.,  Hou B.~L.,  Stickland A.~C.,  Petty J.,  Pang R.~Y.,  Dirani J.,
  Michael J.,   Bowman S.~R.,  2023, GPQA: A Graduate-Level Google-Proof Q\&A
  Benchmark (\mn@eprint {arXiv} {2311.12022})

\bibitem[\protect\citeauthoryear{{Research AI4Science} \& {Azure
  Quantum}}{{Research AI4Science} \& {Azure
  Quantum}}{2023}]{2023arXiv231107361R}
{Research AI4Science} M.,  {Azure Quantum} M.,  2023, arXiv e-prints, \href
  {https://ui.adsabs.harvard.edu/abs/2023arXiv231107361R} {p. arXiv:2311.07361}

\bibitem[\protect\citeauthoryear{Schut, Tomasev, McGrath, Hassabis, Paquet  \&
  Kim}{Schut et~al.}{2023}]{schut2023bridging}
Schut L.,  Tomasev N.,  McGrath T.,  Hassabis D.,  Paquet U.,   Kim B.,  2023,
  Bridging the Human-AI Knowledge Gap: Concept Discovery and Transfer in
  AlphaZero (\mn@eprint {arXiv} {2310.16410})

\bibitem[\protect\citeauthoryear{Steinhardt}{Steinhardt}{2023}]{GPT2030}
Steinhardt J.,  2023, What will GPT-2030 look like?, \url
  {https://bounded-regret.ghost.io/what-will-gpt-2030-look-like/}

\bibitem[\protect\citeauthoryear{Strogatz}{Strogatz}{2000}]{strogatz:2000}
Strogatz S.~H.,  2000, Nonlinear Dynamics and Chaos: With Applications to
  Physics, Biology, Chemistry and Engineering.
Westview Press

\bibitem[\protect\citeauthoryear{{Tamayo}, {Rein}, {Shi}  \&
  {Hernandez}}{{Tamayo} et~al.}{2020}]{2020MNRAS.491.2885T}
{Tamayo} D.,  {Rein} H.,  {Shi} P.,   {Hernandez} D.~M.,  2020, \mn@doi
  [\mnras] {10.1093/mnras/stz2870}, \href
  {https://ui.adsabs.harvard.edu/abs/2020MNRAS.491.2885T} {491, 2885}

\bibitem[\protect\citeauthoryear{Touvron et~al.,}{Touvron
  et~al.}{2023}]{llama2}
Touvron H.,  et~al., 2023, Llama 2: Open Foundation and Fine-Tuned Chat Models
  (\mn@eprint {arXiv} {2307.09288})

\bibitem[\protect\citeauthoryear{{Virtanen} et~al.,}{{Virtanen}
  et~al.}{2020}]{2020NatMe..17..261V}
{Virtanen} P.,  et~al., 2020, \mn@doi [Nature Methods]
  {10.1038/s41592-019-0686-2}, \href
  {https://ui.adsabs.harvard.edu/abs/2020NatMe..17..261V} {17, 261}

\bibitem[\protect\citeauthoryear{Wang et~al.,}{Wang
  et~al.}{2023a}]{wang2023scibench}
Wang X.,  et~al., 2023a, SciBench: Evaluating College-Level Scientific
  Problem-Solving Abilities of Large Language Models (\mn@eprint {arXiv}
  {2307.10635})

\bibitem[\protect\citeauthoryear{{Wang} et~al.,}{{Wang}
  et~al.}{2023b}]{2023arXiv231003731W}
{Wang} K.,  et~al., 2023b, \mn@doi [arXiv e-prints]
  {10.48550/arXiv.2310.03731}, \href
  {https://ui.adsabs.harvard.edu/abs/2023arXiv231003731W} {p. arXiv:2310.03731}

\bibitem[\protect\citeauthoryear{Yang, Li, Lin, Wang, Lin, Liu  \& Wang}{Yang
  et~al.}{2023}]{yang2023dawn}
Yang Z.,  Li L.,  Lin K.,  Wang J.,  Lin C.-C.,  Liu Z.,   Wang L.,  2023, The
  Dawn of LMMs: Preliminary Explorations with GPT-4V(ision) (\mn@eprint {arXiv}
  {2309.17421})

\bibitem[\protect\citeauthoryear{Zhang et~al.,}{Zhang
  et~al.}{2023}]{zhang2023sirens}
Zhang Y.,  et~al., 2023, Siren's Song in the AI Ocean: A Survey on
  Hallucination in Large Language Models (\mn@eprint {arXiv} {2309.01219})

\makeatother
\end{thebibliography}

\appendix


\section{List of problems}
Below is the complete and final list of problems included in our main analysis, grouped by code base. 

\subsection{\texttt{REBOUND}}

\begin{enumerate}
\item 
Use the Rebound N-body integration Python package version 3.12.2 to calculate the minimum mass of exoplanet Wasp-47b needed for exoplanet Wasp-47e's orbit to be unstable on a 100 year timescale, as a function of its eccentricity. Give me a fully working code.

\item 
Use the Rebound N-body integration Python package version 3.12.2 to solve the following problem. Assuming the presence of a fictitious planet orbiting between Jupiter and Neptune, what can its maximum mass be for its orbit to remain stable after 1000 years ? Give me a fully working code.

 \item 
Use the Rebound N-body integration Python package version 3.12.2 to calculate the next date and time when Uranus' mean anomaly is 0 degrees. Give me a fully working code.

\item 
Use the Rebound N-body integration Python package version 3.12.2, and its extension Reboundx version 3.1.0, to solve the following problem. Assuming a Pluto-like object orbiting on a circular coplanar orbit at 25 AU, and Neptune migrating outwards starting at 20 AU, what is the maximum allowed orbital migration timescale of Neptune for this object to be captured in the 3:2 mean motion resonance, as a function of Neptune's eccentricity ? Give me a fully working code.
 \item 

Use the Rebound N-body integration Python package version 3.12.2 to solve the following problem. Assuming the presence of a fictitious planet called HD189733b with Earth-like mean density orbiting between Jupiter and Neptune, what can its maximum radius be for Vesta's orbit to remain stable after 1000 years ? Give me a fully working code.

 \item 
Suppose we replace Callisto with an Earth-mass object. Use the Rebound N-body integration Python package version 3.12.2 to estimate the Lyapunov timescale of the system. Give me a full working code.
 \item 
Use the Rebound N-body integration Python package version 3.12.2 to solve the following problem. Give me step by step instructions to formally check whether a planet is caught in a mean motion resonance, without any prior knowledge about which resonance it might be. Give me a fully working code.
 \item 
Use the Rebound N-body integration Python package version 3.12.2 to solve the following problem. If centaur Chariklo was to double its mass, how far can its rings be while remaining on stable orbits assuming they are made of 2 cm spheres with the same density as Charon ? Give me a fully working code.
 \item 
Use the Rebound N-body integration Python package version 3.12.2, and its extension Reboundx version 3.1.0, to calculate the maximal allowed semimajor axis of Io for its surface volcanic activity to persist, as a function of Io's mean density. Give me a fully working code.
 \item 

Use the Rebound N-body integration Python package version 3.12.2, and its extension Reboundx version 3.1.0, to solve the following problem. Assuming a Pluto-like object orbiting at 25 AU, and Neptune migrating outwards starting at 20 AU, what is the maximum allowed orbital migration timescale of Neptune for this object to be captured in any first order mean motion resonance ? Give me a fully working code.
 \item 
Use the Rebound N-body integration Python package version 3.12.2, and its extension Reboundx version 3.1.0, to solve the following problem. Assume a lunar-mass object with high eccentricity and negative charge orbiting a positively charged Earth-mass object. How small can its orbital period be for the system to be stable for 1000 yrs.
\end{enumerate}

\subsection{\texttt{MESA}}
\begin{enumerate}
\item 
Solve the following problem by providing the complete inputs to the latest version of the MESA stellar physics code (\texttt{inlist}\_project file setting the problem's options and parameters, and run\_star\_extras.f fortran file adding extra physics). You should also provide the python data analysis and postprocessing code needed to generate a complete final solution from MESA's .data output files.
Problem: for what metallicity, if any, will a 3.14159 solar mass star live longer than 10 Gyr ? 
\item 
Solve the following problem by providing the complete inputs to the latest version of the MESA stellar physics code (\texttt{inlist}\_project file setting the problem's options and parameters, and run\_star\_extras.f fortran file adding extra physics). You should also provide the python data analysis and postprocessing code needed to generate a complete final solution from MESA's .data output files.
Problem: what is the expected lifetime of a 30.1861 solar mass star? 
\item 
Solve the following problem by providing the complete inputs to the latest version of the MESA stellar physics code (\texttt{inlist}\_project file setting the problem's options and parameters, and run\_star\_extras.f fortran file adding extra physics). You should also provide the python data analysis and postprocessing code needed to generate a complete final solution from MESA's .data output files.
Problem: what is the expected lifetime of a solar mass star made entirely of helium ? That is, the helium mass fraction at t=0 is 1.

 \item 
Solve the following problem by providing the complete inputs to the latest version of the MESA stellar physics code (\texttt{inlist}\_project file setting the problem's options and parameters, and run\_star\_extras.f fortran file adding extra physics). You should also provide the python data analysis and postprocessing code needed to generate a complete final solution from MESA's .data output files.
Problem: How much energy does a 112.43 solar mass star with 0.5 solar metallicity emit as radiation during the main sequence phase of its evolution?
 \item 
Solve the following problem by providing the complete inputs to the latest version of the MESA stellar physics code (\texttt{inlist}\_project file setting the problem's options and parameters, and run\_star\_extras.f fortran file adding extra physics). You should also provide the python data analysis and postprocessing code needed to generate a complete final solution from MESA's .data output files.
Problem: for a 2.76 solar mass star with non-solar metallicity, at what age does its core helium mass reach exactly 3 times 0.2 solar masses ?
 \item 
Solve the following problem by providing the complete inputs to the latest version of the MESA stellar physics code (\texttt{inlist}\_project file setting the problem's options and parameters, and run\_star\_extras.f fortran file adding extra physics). You should also provide the python data analysis and postprocessing code needed to generate a complete final solution from MESA's .data output files.
Problem: at what time (in Myr) during its evolution will a 30.1861 solar mass star undergo carbon burning ?
 \item 
Solve the following problem by providing the complete inputs to the latest version of the MESA stellar physics code (\texttt{inlist}\_project file setting the problem's options and parameters, and run\_star\_extras.f fortran file adding extra physics). You should also provide the python data analysis and postprocessing code needed to generate a complete final solution from MESA's .data output files.
Problem: what is the expected lifetime of a solar mass and metallicity star made entirely of helium ? That is, the helium mass fraction at t=0 is 1. Assume that G constant is a twice its real value

 \item 
Solve the following problem by providing the complete inputs to the latest version of the MESA stellar physics code (\texttt{inlist}\_project file setting the problem's options and parameters, and run\_star\_extras.f fortran file adding extra physics). You should also provide the python data analysis and postprocessing code needed to generate a complete final solution from MESA's .data output files.
Problem: what is the convective mass (that is the total mass of the gas in the convection zones) of a 3.14 solar mass and solar metallicity star?
 \item 
Solve the following problem by providing the complete inputs to the latest version of the MESA stellar physics code (\texttt{inlist}\_project file setting the problem's options and parameters, and run\_star\_extras.f fortran file adding extra physics). You should also provide the python data analysis and postprocessing code needed to generate a complete final solution from MESA's .data output files.
Problem: for what range of metallicities, if any, will a 26.1451 solar mass main sequence star have 2 separate convective zones ?
 \item 
Solve the following problem by providing the complete inputs to the latest version of the MESA stellar physics code (\texttt{inlist}\_project file setting the problem's options and parameters, and run\_star\_extras.f fortran file adding extra physics). You should also provide the python data analysis and postprocessing code needed to generate a complete final solution from MESA's .data output files.
Problem: what should be the minimal mass of a solar metallicity star for it to completely disperse a 100 solar masses nebula around it through stellar winds ?

 \item 
Solve the following problem by providing the complete inputs to the latest version of the MESA stellar physics code (\texttt{inlist}\_project file setting the problem's options and parameters, and run\_star\_extras.f fortran file adding extra physics). You should also provide the python data analysis and postprocessing code needed to generate a complete final solution from MESA's .data output files.
Problem: what is the main sequence lifetime of a 254.54 solar masses star that has a 4.23 solar mass and 10 solar luminosities black hole at its core ? Provide the complete MESA code \texttt{inlist}\_project file and python data analysis and postprocessing code that when run provide the precise answer.

\end{enumerate}

\subsection{\texttt{Dedalus}}
\begin{enumerate}

 \item 
Consider a one-dimensional diffusion problem in the spatial domain [0,1]. The initial condition is a domain-centred Gaussian of Full-Width-at-Half-Maximum 0.3 with unit peak amplitude. The boundary conditions are outflow on each side. The diffusion coefficient is constant: D=0.1. When does the peak value in the domain reach 10\% of its initial peak value? Use the Dedalus PDE solver Python package, version 2. Give me a fully working code.
 \item 
Consider a one-dimensional scalar advection problem in the spatial domain [0,1]. The advection flow is one-directional and has positive constant velocity v=0.1. The boundary conditions are outflow on each side. A domain-centred Gaussian scalar source of Full-Width-at-Half-Maximum 0.3 with peak rate amplitude S=0.01 is turned on at t=0. Assuming a uniform scalar field initial condition, with unit value, how long does it take for the system to reach steady state? Use the Dedalus PDE solver Python package, version 2. Give me a fully working code.
 \item 
Consider a one-dimensional scalar advection problem in the spatial domain [0,1]. The advection flow is one-directional and has positive constant velocity v=0.1. The boundary conditions are outflow on each side. A domain-centred Gaussian scalar source of Full-Width-at-Half-Maximum 0.3 with peak rate amplitude S=0.01 is turned on at t=0. Assuming a uniform scalar field initial condition, with unit value, what is the maximum scalar value over the domain once the system has reached steady state? Use the Dedalus PDE solver Python package, version 2. Give me a fully working code.
 \item 
Consider a non-linear one-dimensional acoustic wave problem in the spatial domain x=[0,1]. The continuous variable is u and the wave velocity c(u) = u\^a is a polynomial function of the local field amplitude u, with non-negative exponent a. The velocity exponent a is set to unity for this problem. The boundary conditions are periodic. The initial condition is a domain-centred Gaussian u-field of Full- Width-at-Half-Maximum 0.3 with unit peak amplitude. How long does it take for the wave crest to first reach the x=1 boundary? Use the Dedalus PDE solver Python package, version 2. Give me a fully working code.
 \item 
Consider a one-dimensional diffusion problem in the spatial domain [0,1]. The initial condition is a domain-centred Gaussian of Full-Width-at-Half-Maximum 0.3 of arbitrary amplitude. The boundary conditions are outflow on each side. The diffusion coefficient is constant with value D\_0. How much larger than the reference value D\_0 does the diffusion coefficient D need to be for the peak value in the domain to reach 10\% of the initial peak value three times faster than in the case D=D\_0? Use the Dedalus PDE solver Python package, version 2. Give me a fully working code.
 \item
Consider a one-dimensional scalar advection problem in the spatial domain [0,1]. The advection flow is one-directional and has positive constant velocity v=0.1. The boundary conditions are outflow on each side. A domain-centred Gaussian scalar source of Full-Width-at-Half-Maximum 0.3 with peak rate amplitude S=0.01 is turned on at t=0. Assuming a uniform scalar field initial condition, with unit value, how much stronger should the source rate S be for the steady-state to be reached twice as fast? Use the Dedalus PDE solver Python package, version 2. Give me a fully working code.

 \item 
Consider a one-dimensional diffusion problem in the spatial domain x=[0,1]. The initial condition is a domain-centred Gaussian of Full-Width-at-Half-Maximum 0.3 with unit peak amplitude. The boundary conditions are outflow on each side. The diffusion coefficient is spatially inhomogeneous: it increases linearly with x, from value D=1 in x=0 to value D=2 in x=1 . After how much time does the peak value in the domain reach 10\% of the initial peak value? Use the Dedalus PDE solver Python package, version 2. Give me a fully working code.
 \item 
Consider a one-dimensional scalar advection problem in the spatial domain [0,1]. The advection flow is one-directional and has positive constant velocity v=0.1. The boundary conditions are outflow on each side. A domain-centred Gaussian scalar source of Full-Width-at-Half-Maximum 0.3 with peak rate amplitude S=0.01 is turned on at t=0. A spatially and temporally uniform sink is also turned on at t=0. Assuming a uniform scalar field initial condition, with unit value, how long does it take for the system to reach steady state? Use the Dedalus PDE solver Python package, version 2. Give me a fully working code.
 \item 
Consider a non-linear one-dimensional acoustic wave problem in the spatial domain x=[0,1]. The continuous variable is u and the wave velocity c(u) = u\^a is a polynomial function of the local wave amplitude u, with non-negative exponent a. The velocity exponent a is set to unity for this problem . The boundary conditions are periodic. The initial condition is a domain-centred Gaussian u-field of Full- Width-at-Half-Maximum 0.3 with unit peak amplitude. How long does it take for the system to first develop a shock-wave? Use the Dedalus PDE solver Python package, version 2. Give me a fully working code.
 \item 
Consider a one-dimensional diffusion problem in the spatial domain x=[0,1]. The initial condition is a domain-centred Gaussian of Full-Width-at-Half-Maximum 0.3 of arbitrary amplitude. The boundary conditions are outflow on each side. The diffusion coefficient is spatially inhomogeneous: it increases linearly with x from value D\_0=1 in x=0 to the value D\_0=2 in x=1. By how much should one uniformly increase D over the reference profile D\_0 for the peak value in the domain to reach 10\% of the initial peak value three times faster than for the case of the reference profile D\_0? Use the Dedalus PDE solver Python package, version 2. Give me a fully working code.

\item 
Consider a one-dimensional scalar advection problem in the spatial domain [0,1]. The advection flow is one-directional and has positive constant velocity v=0.1. The boundary conditions are outflow on each side. A domain-centred Gaussian scalar source of Full-Width-at-Half-Maximum 0.3 with peak rate amplitude S=0.01 is turned on at t=0. A spatially and temporally uniform sink is also turned on at t=0. Assuming a uniform scalar field initial condition, with unit value, how much stronger should the source rate be, keeping the sink rate unchanged, for the steady-state to be reached twice as fast? Use the Dedalus PDE solver Python package, version 2. Give me a fully working code.
 \item 
Consider a one-dimensional scalar advection problem in the spatial domain [0,1]. The advection flow is one-directional and has positive constant velocity v=0.1. The boundary conditions are outflow on each side. A domain-centred Gaussian scalar source of Full-Width-at-Half-Maximum 0.3 with peak rate amplitude S=0.01 is turned on at t=0. A spatially and temporally uniform sink is also turned on at t=0. Assuming a uniform scalar field initial condition, with unit value, how much weaker should the sink rate be, keeping the source rate unchanged, for the steady-state to be reached twice as fast? Use the Dedalus PDE solver Python package, version 2. Give me a fully working code.
 \item 
Consider a non-linear one-dimensional acoustic wave problem in the spatial domain x=[0,1]. The continuous variable is u and the wave velocity c(u) = u\^a is a polynomial function of the local wave amplitude u, with non-negative exponent a. The boundary conditions are periodic. The initial condition is a domain-centred Gaussian u-field of Full-Width-at-Half-Maximum 0.3 with unit peak amplitude. How weak should the exponent a be for the first shock-wave to develop between the 3rd and the 4th passage of the wave crest across the domain? Use the Dedalus PDE solver Python package, version 2. Give me a fully working code.
 \item 
Consider a one-dimensional diffusion problem in the spatial domain [0,1]. The initial condition is a domain-centred Gaussian of Full-Width-at-Half-Maximum 0.3 with unity peak amplitude. The boundary conditions are outflow. The diffusion coefficient is constant. At any given time t$>$0, does the corresponding 4th-order hyperdiffusion problem diffuse more or less of the field out of the domain than the standard 2nd order diffusion problem? Use the Dedalus PDE solver Python package, version 2. Give me a fully working code.
 \item 

Consider a one-dimensional diffusion problem in the spatial domain [0,1]. The initial condition is a domain-centred Gaussian of Full-Width-at-Half-Maximum 0.3 with unity peak amplitude. The boundary conditions are periodic. The diffusion coefficient is constant. Can the peak value in the domain ever reach 10\% of the initial peak value? Use the Dedalus PDE solver Python package, version 2. Give me a fully working code.

 \item 
Consider a one-dimensional diffusion problem in the spatial domain x=[0,1]. The initial condition is a domain-centred Gaussian of Full-Width-at-Half-Maximum 0.3 with unit peak amplitude. The boundary conditions are outflow on each side. The diffusion coefficient is spatially inhomogeneous: it increases linearly with x from value D=1 in x=0 to value D=2 in x=1 . When does the domain-integral of the continuous field being diffused reach 1\% of the initial integral value? Use the Dedalus PDE solver Python package, version 2. Give me a fully working code.
 \item 
Consider a one-dimensional diffusion problem in the spatial domain x=[0,1]. The initial condition is a domain-centred Gaussian of Full-Width-at-Half-Maximum 0.3 of arbitrary amplitude. The boundary conditions are outflow. The diffusion coefficient is spatially inhomogeneous: it increases linearly with x from value D in x=0 to 2D in x=1. How much larger than the reference value D\_0 does D need to be for the domain-integral of the field being diffused to reach 10\% of the initial peak integral value at least three times faster than for the case D=D\_0? Use the Dedalus PDE solver Python package, version 2. Give me a fully working code.
 \item 
Can you build a one-dimensional antidiffusion PDE system in the spatial domain x=[0,1] such that anti- diffusion system reaches steady-state with a (possibly inhomogeneous) constant loss term in the domain? Assume a constant field for the initial condition and outflow boundary conditions. Use the Dedalus PDE solver Python package, version 2. Give me a fully working code.
 \item 
Consider a non-linear one-dimensional acoustic wave problem in the spatial domain x=[0,1]. The continuous variable is u. The wave velocity c(u,t) = u\^a exp(- lambda t) is a polynomial function of the
local wave amplitude u, with non-negative exponent a=1, that is also decaying as a function of time at the decay rate lambda. The boundary conditions are periodic. The initial condition is a domain-centred Gaussian u-field of Full-Width-at-Half-Maximum 0.3 with unit peak amplitude. How strong should the decay rate lambda be for the first shock-wave to develop after the 5th passage of the wave crest across the domain? Use the Dedalus PDE solver Python package, version 2. Give me a fully working code.
 \item 
Consider a non-linear one-dimensional acoustic wave problem in the spatial domain x=[0,1]. The continuous variable is u. The wave velocity c(u,t) = u\^a exp(- lambda t) is a polynomial function of the local wave amplitude u, with non-negative exponent a=1, that is also decaying as a function of time at the decay rate lambda. The boundary conditions are periodic. The initial condition is a domain-centred Gaussian u-field of Full-Width-at-Half-Maximum 0.3 with unit peak amplitude. How strong should the decay rate lambda be to avoid the formation of a shock-wave at any future time. Use the Dedalus PDE solver Python package, version 2. Give me a fully working code.
\end{enumerate}

\subsection{\texttt{Scipy}}
\begin{enumerate}

 \item 
Consider the classical Lorenz dynamical system, with a specific initial condition on the z axis with z=25. What is the time it takes for the system to cross twice the y-z plane? Use the ODE integrator from the Scipy python package. Give me a fully working code.
 \item 
Consider the classical Lorenz dynamical system, with an arbitrary initial condition on the z axis with z=z\_0. What is the value of z\_0 that will minimize the time it takes for the system to cross twice the y-z plane? Use the ODE integrator from the Scipy python package. Give me a fully working code.

 \item 
Consider the classical Lorenz dynamical system, with two specific initial conditions on the z axis, with z=25 and z=23. What is the time it takes for the euclidian distance between these two configuration points to exceed 20? Use the ODE integrator from the Scipy python package. Give me a fully working code.

 \item 
Consider the classical Lorenz dynamical system, except that the sigma parameter exhibits a time- dependent sinusoidal oscillation around the standard value, with an amplitude A=5 and an angular frequency omega=0.1. Use a specific initial condition on the z axis, with z=25. What is the time it takes for the system to cross twice the y-z plane? Use the ODE integrator from the Scipy python package. Give me a fully working code.
 \item 
Consider the classical Lorenz dynamical system, except that the sigma parameter exhibit a time- dependent sinusoidal oscillation around the standard value, with an amplitude A=5 and an angular frequency omega=0.1. Use a specific initial condition on the z axis with z=25. What are the values of A and Omega that will minimize the time it takes for the system to cross twice the y-z plane? Use the ODE integrator from the Scipy python package. Give me a fully working code.

\end{enumerate}

\section{Complete output examples}
\label{outputexamples}
In this section we show representative complete examples for the output of \texttt{GPT4} to our problems. 

\subsection{REBOUND}
Prompt:\\
\noindent\fbox{%
    \parbox{\textwidth}{%
Use the Rebound N-body integration Python package version 3.12.2 to solve the following problem. Assuming the presence of a fictitious planet called HD189733b with Earth-like mean density orbiting between Jupiter and Neptune, what can its maximum radius be for Vesta's orbit to remain stable after 1000 years ? Give me a fully working code.
}}

Unmodified, complete output of LLM:\\
\noindent\fbox{%
    \parbox{\textwidth}{%
 To solve this problem, we will use the Rebound N-body integration package to simulate the Solar System with the addition of the fictitious planet HD189733b. We will then vary the radius of HD189733b and check the stability of Vesta's orbit after 1000 years. Here's the code:

 $<$...$>$

 This code sets up a simulation of the Solar System with the Sun, Jupiter, Neptune, Vesta, and the fictitious planet HD189733b. The radius of HD189733b is varied, and the stability of Vesta's orbit is checked after 1000 years. The maximum radius of HD189733b for Vesta's orbit to remain stable is then printed.
}}

\begin{mintedbox}[]{python}
import numpy as np
import rebound

def setup_simulation(radius):
    """
    Set up the simulation with the Solar System and the fictitious planet HD189733b.
    :param radius: The radius of HD189733b in Earth radii.
    :return: A Rebound simulation object.
    """
    sim = rebound.Simulation()

    # Add the Sun, Jupiter, and Neptune
    sim.add("Sun")
    sim.add("Jupiter")
    sim.add("Neptune")

    # Add Vesta
    sim.add("Vesta")

    # Add HD189733b with Earth-like mean density
    mass = 4/3 * np.pi * radius**3 * 5.52  # Mass in Earth masses
    sim.add(m=mass/333000, a=(5.2+30.1)/2)  # Orbit between Jupiter and Neptune

    # Set the integrator and timestep
    sim.integrator = "ias15"
    sim.dt = sim.particles[1].P * 0.05  # 5% of Jupiter's orbital period

    # Move to the center of mass frame
    sim.move_to_com()

    return sim

def check_vesta_stability(radius, years=1000):
    """
    Check the stability of Vesta's orbit after a given number of years.
    :param radius: The radius of HD189733b in Earth radii.
    :param years: The number of years to simulate.
    :return: True if Vesta's orbit is stable, False otherwise.
    """
    sim = setup_simulation(radius)
    initial_distance = np.linalg.norm(np.array(sim.particles[3].xyz) - np.array(sim.particles[0].xyz))
    sim.integrate(years * 2 * np.pi)

    final_distance = np.linalg.norm(np.array(sim.particles[3].xyz) - np.array(sim.particles[0].xyz))
    return abs(final_distance - initial_distance) < 0.1 * initial_distance

def find_maximum_radius():
    """
    Find the maximum radius of HD189733b for Vesta's orbit to remain stable after 1000 years.
    :return: The maximum radius in Earth radii.
    """
    radius = 1
    while check_vesta_stability(radius):
        radius += 1
    return radius - 1

if __name__ == "__main__":
    max_radius = find_maximum_radius()
    print(f"The maximum radius of HD189733b for Vesta's orbit to remain stable after 1000 years is {max_radius} Earth radii.")
\end{mintedbox}

In this example, we asked \texttt{GPT4} to estimate the stability timescale of the solar system's minor planet Vesta as a function of (explicitly) the radius and (implicitly) the location of a hypothetical ninth planet named (confusingly by design) after exoplanet HD189733b. \texttt{GPT4}'s solution was graded  Pass- by the authors, even though it gives the wrong answer. Moreover, the following was noted:
\begin{itemize}
    \item The code runs with no errors. 
    \item However, the main issue stopping the code from giving a correct answer is units conversion and physical consistency. \texttt{GPT4} defined Vesta's radius in unit-less values with the order of 1 (R=1,2,3.. with no explicit units), then proceeds to calculates the mass while taking Earth's density as 5.5 g/cm$^3$. The radius is hence in cm, and using order 1 values does not make sense (Vesta's radius is 26.2 million cm). Moreover, \texttt{GPT4} finally declared the mass to be in Earth masses, even though it is actually in grams, and proceeded to convert it to solar masses. 
    \item \texttt{GPT4} did otherwise deal correctly with the G=1 default REBOUND units. 
    \item As a measure of stability, \texttt{GPT4} used the distance of Vesta to the sun. It considered the orbit to be unstable if it varied by more than 10\% over the 1000 yr timescale. While this definition of stability is neither formal nor particularly good, we judged it to be acceptable for a passing grade. 
    \item \texttt{GPT4} calculated the distance using Vesta's Cartesian coordinates instead of just relying on the values readily available in REBOUND.
\end{itemize}

\subsection{MESA}

Prompt:\\
\noindent\fbox{%
    \parbox{\textwidth}{%
Solve the following problem by providing the complete inputs to the latest version of the MESA stellar physics code (inlist\_project file setting the problem's options and parameters, and run\_star\_extras.f fortran file adding extra physics). You should also provide the python data analysis and postprocessing code needed to generate a complete final solution from MESA's .data output files.
Problem: at what time (in Myr) during its evolution will a 30.1861 solar mass star undergo carbon burning ?
}}

Unmodified, complete output of LLM:\\

\noindent\fbox{%
    \parbox{\textwidth}{%
  As a language model AI, I cannot run the MESA code directly, but I can provide you with the necessary input files and Python code to analyze the output data. You can then run the MESA code on your local machine or server.
  $<$...$>$
  This Python script uses the ``mesa\_reader'' package to read the MESA output data. Make sure to install it using ``pip install mesa\_reader'' before running the script.

After running the Python script, you will get the time (in Myr) when the 30.1861 solar mass star undergoes carbon burning.
}}

\begin{mintedbox}[]{python}


First, create the `inlist_project` file with the following content:

```
! inlist_project
! Input file for MESA stellar evolution code
! Problem: 30.1861 solar mass star undergoing carbon burning

&star_job
  ! Set the initial mass of the star
  initial_mass = 30.1861
  ! Save the output data
  save_model_when_terminate = .true.
  ! Set the output directory
  output_directory = '30.1861Msun_star'
/

&controls
  ! Set the initial metallicity
  initial_z = 0.02
  ! Set the mixing length theory parameter
  mixing_length_alpha = 1.5
  ! Set the maximum number of steps
  max_num_steps = 10000
  ! Set the maximum number of retries
  max_num_retries = 100
  ! Set the maximum number of backups
  max_num_backups = 100
  ! Set the stopping condition for carbon burning
  stop_at_carbon_burning = .true.
/

&pgstar
  ! Set the output file format
  file_format = 'png'
  ! Set the output directory
  output_directory = '30.1861Msun_star'
/
```

Next, create the `run_star_extras.f` file with the following content:

```fortran
! run_star_extras.f
! Fortran file for adding extra physics to MESA code
! Problem: 30.1861 solar mass star undergoing carbon burning

! No extra physics needed for this problem
```

Now, run the MESA code with these input files. After the code finishes running, you will have a set of `.data` output files in the `30.1861Msun_star` directory.

Finally, create a Python script to analyze the output data and find the time when the star undergoes carbon burning:

```python
import os
import numpy as np
import matplotlib.pyplot as plt
from mesa_reader import MesaData

def find_carbon_burning_time(output_directory):
    # Find the last .data file in the output directory
    data_files = sorted([f for f in os.listdir(output_directory) if f.endswith('.data')])
    last_data_file = os.path.join(output_directory, data_files[-1])

    # Read the data from the last .data file
    data = MesaData(last_data_file)

    # Find the time when the star undergoes carbon burning (in Myr)
    carbon_burning_time = data.star_age / 1e6

    return carbon_burning_time

output_directory = '30.1861Msun_star'
carbon_burning_time = find_carbon_burning_time(output_directory)
print(f"The 30.1861 solar mass star undergoes carbon burning at {carbon_burning_time:.2f} Myr.")
```


\end{mintedbox}

In this example, we asked \texttt{GPT4} to estimate when would a star with a given (floating point value) mass undergo carbon burning during its post main sequence evolution. An implicit parameter is the star's metallicity. The LLM failed at this exercice since it hallucinated a stopping condition ``stop\_at\_carbon\_burning'' and took the stopping time as the solution. As no such stopping condition exists in MESA, acceptable solutions would have either manually implemented such a condition in the Fortran code, or let the simulation run through to the end then searched for carbon burning point in the data post-processing.  Moreover, the following was noted:
\begin{itemize}
    \item There are many coding errors in the \texttt{Inlist}, mainly flags that in reality exist under different name. 
    \item On a physics note, the LLM automatically enabled set the mixing length parameter (mixing\_length\_alpha) to an acceptable value (1.5).
    \item Using floating point and confusing stellar masses never had any influence on the results. 
\end{itemize}

\subsection{Dedalus}

\noindent\fbox{%
    \parbox{\textwidth}{%
    Prompt: Consider a one-dimensional diffusion problem in the spatial domain [0,1]. The initial condition is a domain-centred Gaussian of Full-Width-at- Half-Maximum 0.3 with unit peak amplitude. The boundary conditions are outflow on each side. The diffusion coefficient is constant: D=0.1. When does the peak value in the domain reach 10\% of its initial peak value? Use the Dedalus PDE solver Python package, version 2. Give me a fully working code.
}}

~\newline
Output of LLM: To solve this problem, we will use the Dedalus PDE solver package to simulate the one-dimensional diffusion problem with the given initial and boundary conditions. We will then find the time when the peak value in the domain reaches 10\% of its initial peak value. Here's the fully working code:

\begin{mintedbox}[]{python}
import numpy as np
import matplotlib.pyplot as plt
from dedalus import public as de 
from dedalus.extras import flow_tools 
import time

# Constants
FWHM = 0.3
D = 0.1 
tolerance = 0.1

# Create the domain
x_basis = de.Chebyshev('x', 256, interval=(0, 1)) 
domain = de.Domain([x_basis], grid_dtype=np.float64)

# Define the problem
problem = de.IVP(domain, variables=['u', 'ux'])
problem.parameters['D'] = D 
problem.add_equation("dt(u) - D*dx(ux) = 0") 
problem.add_equation("ux - dx(u) = 0") 
problem.add_bc("left(ux) = 0") 
problem.add_bc("right(ux) = 0")

# Set the initial condition
x = domain.grid(0)
u = domain.new_field(name='u')
u['g'] = np.exp(-4 * np.log(2) * (x - 0.5)**2 / FWHM**2)

# Set up the solver
solver = problem.build_solver(de.timesteppers.RK443) 
solver.stop_sim_time = np.inf
solver.stop_wall_time = 30 * 60
solver.stop_iteration = np.inf

# Set up the analysis
analysis = solver.evaluator.add_file_handler('data', iter=10)
analysis.add_system(solver.state, layout='g')
analysis.add_task("integ(u,'x')", layout='g', name='total_mass')

# Main loop
initial_peak_value = np.max(u['g']) 
target_peak_value = 0.1 * initial_peak_value 
start_time = time.time()

while solver.ok:
    solver.step(dt=1e-3)
    current_peak_value = np.max(solver.state['u']['g'])
    if current_peak_value <= target_peak_value: 
        break

end_time = time.time()
print("Simulation completed in {:.2f} seconds.".format(end_time - start_time))
print("The peak value reaches 10\% of its initial value at t = {:.5f}.".format(solver.sim_time))

# Plot the final state
plt.plot(x, solver.state['u']['g'], label='Final state') plt.xlabel('x')
plt.ylabel('u')
plt.legend()
plt.show()
\end{mintedbox}

In this example, we asked GPT4 to setup and solve a 1D diffusion problem to evaluate when the peak value in the domain reaches 10\% of the peak initial value, as the original gaussian peak diffuses out through the boundaries. The code is largely satisfactory in setting up the domain, the equations, boundary and initial conditions, time-stepping and the requested diagnostic.

This code solution gets a Pass+ grade. Three ``analysis'' lines are unnecessary (and unused) but it illustrates satisfactory performance on one of our simpler diffusion problems.

\subsection{Scipy}

\noindent\fbox{%
    \parbox{\textwidth}{%
    Prompt: Consider the classical Lorenz dynamical system, with two specific initial conditions on the z axis, with z=25 and z=23. What is the time it takes for the euclidian distance between these two configuration points to exceed 20? Use the ODE integrator from the Scipy python package. Give me a fully working code.
}}
\newline

Output of LLM: Here's a fully working Python code that uses the Scipy package to solve the Lorenz dynamical system and find the time it takes for the Euclidean distance between the two configuration points to exceed 20:

\begin{mintedbox}[]{python}
import numpy as np
from scipy.integrate import solve_ivp 
import matplotlib.pyplot as plt

# Define the Lorenz system of equations
def lorenz_system(t, xyz, sigma=10, rho=28, beta=8/3): 
    x, y, z = xyz
    dx_dt = sigma * (y - x) 
    dy_dt = x * (rho - z) - y 
    dz_dt = x * y - beta * z 
    return [dx_dt, dy_dt, dz_dt]
    
# Define the stopping condition when the Euclidean distance exceeds 20
def stop_condition(t, xyz1_xyz2):
    xyz1, xyz2 = np.split(xyz1_xyz2, 2) 
    distance = np.linalg.norm(xyz1 - xyz2) 
    return distance - 20
    
stop_condition.terminal = True

# Initial conditions
xyz1_initial = [1, 1, 25] 
xyz2_initial = [1, 1, 23]

# Combine the initial conditions for both points
initial_conditions = np.concatenate((xyz1_initial, xyz2_initial)) 

# Integrate the Lorenz system using solve_ivp
solution = solve_ivp(lorenz_system, (0, 100), initial_conditions, events=stop_condition, dense_output=True) 

# Extract the time when the stopping condition is met
time_to_exceed_20 = solution.t_events[0][0]
print(f"The time it takes for the Euclidean distance between the two configuration points to exceed 20 is: {time_to_exceed_20:.2f}")

# Plot the trajectories of the two points in 3D
fig = plt.figure()
ax = fig.add_subplot(111, projection="3d")
t = np.linspace(0, time_to_exceed_20, 1000) 
xyz1_xyz2 = solution.sol(t)
xyz1, xyz2 = np.split(xyz1_xyz2, 2)
ax.plot(xyz1[0], xyz1[1], xyz1[2], label="z=25") 
ax.plot(xyz2[0], xyz2[1], xyz2[2], label="z=23")
ax.set_xlabel("X") 
ax.set_ylabel("Y") 
ax.set_zlabel("Z") 
ax.legend()
plt.show()



\end{mintedbox}

In this example, we asked GPT4 to evaluate two distinct initial conditions diverge in time for the standard Lorenz system, a defining property of deterministic chaos. This code solution gets a Fail grade because of several compounding errors. The initial conditions chosen by GPT4 do not follow prompt specifications, the concatenation of trajectory variables is problematic and the simulation time is arbitrary. 

Note that GPT4 sets up a correct and modular Lorenz system function and uses a compact algorithmic approach to detect when the Euclidian distance reaches the desired threshold.

\section{Exploration}
\label{app:explore}

\subsection{LLM choice \& comparison}
During our initial exploration phase (detailed in section \ref{calibration}), we considered Llama2-70B-Chat \citep{llama2}, Falcon-180B-chat\footnote{https://falconllm.tii.ae/}, Claude2\footnote{https://www.anthropic.com/index/claude-2} and GPT4\footnote{https://openai.com/research/gpt-4} as candidate LLMs for our work. A brief survey of capabilities in computational physics indicated that LLama-2 models are not sufficiently competent for the task. Claude2 and Falcon-180B showed noticeably weaker performance than GPT4. As a result, we focused on \texttt{GPT4} for the remainder of our work.

\subsection{GPT4 exploration}
\label{calibration}
Before proceeding to systematically benchmark GPT4's performance on our $\sim 50$ original problems, we explored \texttt{GPT4} capabilities in computational physics to help us calibrate our methods and prompting strategy. We performed this exploration using a code base and physics problems that are distinct from our benchmark problems, to avoid any obvious design contamination/leakage.   This exploration helped us identify some of \texttt{GPT4}'s failure modes and ultimately guided us towards our simple prompting strategy (\S~\ref{sec:prompt}).

We created multiple problems of varying difficulty to be solved using the Python package for galactic dynamics \texttt{Galpy}\footnote{https://github.com/jobovy/galpy} \citep{galpy1,galpy2}. This package offers capabilities such as orbit integration in diverse potentials, assessment and sampling of different distribution functions, and computing action-angle coordinates in all static potentials.

While still being in the general domain of gravitational dynamics, \texttt{Galpy} is different from \texttt{REBOUND} in that it is not a general N-body integrator. It does share with  \texttt{REBOUND} an applicability in the general context of planetary and stellar dynamics and contains many analogous physics concepts and tools.  

The exploration with \texttt{GPT4} (\texttt{gpt-4-0314}, with \texttt{temperature}=0) was done by asking a series of questions, in a multi-turn dialogue setup, starting with the basic prompt template:

\noindent\fbox{%
    \parbox{\textwidth}{%
You are a useful coding assistant for a scientist. You help solve specific physics problem by writing python code that makes use of domain-specialized python packages, as requested by the scientist. Your code is thoroughly commented for clarity and to provide descriptions of the various physics and coding choices made in your solution. You should always follow precisely the given instructions without omissions. Use Galpy 1.7.0 to $<$...$>$. Give me full working code.
    }%
}

We give here a quick overview of the topics and questions covered during the exploration, and the lessons learned from it. The detailed prompts and LLM answers can be found in Appendix~D. Three sets of problems were discussed with \texttt{GPT4}:
\begin{itemize}
    \item Orbital stability of stars in different galactic potentials. The LLM was able to define a 
    reasonable (although ultimately scientifically inaccurate for the context) stability criterion, and provide a working \texttt{Galpy} code. Followup questions where asked concerning the effects of different potentials, and the presence of black holes. 
    \item Orbital integration of Earth in a Keplerian potential. The LLM provided a reasonable and working \texttt{Galpy} code.
    \item Orbital integration of the 8 solar system planets in a 3D Keplerian potential. The LLM struggled significantly, and a very detailed prompt using chain-of-thoughts and self-critic elements was required for the LLM to give a reasonable answer. We interpreted this as evidence of the LLM struggling with out-of-distribution questions, since integrating the solar system is not what \texttt{Galpy} was designed for, nor is it usually used for that (so few online examples to learn from must exist). 
\end{itemize}

The lessons we learned from this exploration are listed here (and referenced in Appendix~D):

\begin{enumerate}
\item \textit{Lesson 1: }It is important to specify the version of the code one wants the LLM to use, and to make sure that this version precedes the ``training cutoff date'' beyond which the LLM has no knowledge. For the version of \texttt{GPT4} we use, that would be September 2021. 
 \item\textit{Lesson 2: } One cannot always trust the physics interpretations. While most of the dialogue might seem and be reasonable, serious conceptual flaws can emerge (e.g., claiming that \texttt{NFWPotential} has the longest stability timescale, instead of the shortest, even though the (correct) opposite was stated by the same LLM earlier in the conversation).
\item \textit{Lesson 3: } One benefits from always explicitly prompting the LLM to implement error handling for edge cases. 
\item \textit{Lesson 4:} \texttt{GPT4} can capably discuss, generate code, and offer reasonable interpretations for some slightly out-of-distribution scenarios. 
\item \textit{Lesson 5:} \texttt{GPT4} can perform surprisingly poorly at visualizing numerical results. The plotting request should be very detailed, explicitly mentioning whether to use built-in functions or matplotlib (much preferred), and the library version. 
\item \textit{Lesson 6:} \texttt{GPT4} can occasionally self-reflect on some physics problems and offer research-grade alternatives. 
\item \textit{Lesson 7:} \texttt{GPT4} will often implement technically correct but overly-simplified implementations of the requested solution. Followup prompting in a multi-turn dialogue setup can help develop the code further.

\end{enumerate}

\section{Details on Exploration}
\textcolor{black}{Below we provide details of our exploration of GPT4's capabilities using \texttt{Galpy}, which served as calibration for the problem design methods described in the main text. Note that, in the following, 3 types of text are shown: (i) prompts used to condition the LLM generation (all shown inside text boxes, starting with ``Prompt:''), (ii) GPT4 responses (either inside a text box starting with ``GPT4:'', or inside a code box to facilitate code) readability, and (iii) finally labeled Exploration Notes by the authors that are not part of the dialogue with the LLM, are \textit{not} inside text boxes and are all highlighted in \textit{italics}. Furthermore, when the starting sentence is ``Followup prompt'', this implies that prompt is part of a multi-turn dialogue, while a plain ``Prompt'' implies the first round of a new dialogue. In general,  we started dialogues anew whenever a new topic or problem was introduced.  }

\subsubsection{Stellar orbit stability}

\noindent\fbox{%
    \parbox{\textwidth}{%
Prompt: Use Galpy 1.7.0 to estimate the stability timescale of a massive star orbiting at 20 pc from the center of the milkyway as a function of the gravitational potential used. Give me full working code.
    }%
}

\textit{Exploration note: relates to Lesson 1 in Appendix~C.}

\noindent\fbox{%
    \parbox{\textwidth}{%
GPT4: Sure, I can help with that. Here's a Python script that uses the galpy package to estimate the stability timescale of a massive star orbiting at 20 parsecs from the center of the Milky Way. This script uses three different gravitational potentials: the MWPotential2014, NFWPotential, and HernquistPotential. This script assumes that the star is initially at rest relative to the galactic center. The stability timescale is estimated as the time it takes for the star's radial distance from the galactic center to change by 10\%. This is a rough estimate and the actual stability timescale could be longer or shorter depending on the specific dynamics of the system.

Please note that the galpy package uses natural units where the gravitational constant G = 1, the total mass of the Milky Way M = 1, and the characteristic length scale a = 1. The potentials are normalized so that the total mass equals 1. The NFWPotential and HernquistPotential are parameterized by a scale length a, which is set to 1 in this script. You may need to adjust these parameters depending on the specific properties of the Milky Way.
    }%
}

\begin{mintedbox}[]{python}
# Import necessary modules
import numpy as np
from galpy.potential import MWPotential2014, NFWPotential, HernquistPotential
from galpy.orbit import Orbit
from galpy.util.bovy_conversion import time_in_Gyr
from astropy import units

# Define the initial conditions of the star
# Assuming the star is at rest relative to the galactic center
R = 20. * units.pc.to(units.kpc)  # Convert parsecs to kiloparsecs
vR, vT, vz = 0., 0., 0.  # Initial velocities
phi = 0.  # Initial azimuth

# Define the potentials
potentials = [MWPotential2014, NFWPotential(a=1., normalize=1.), HernquistPotential(a=1., normalize=1.)]
potential_names = ['MWPotential2014', 'NFWPotential', 'HernquistPotential']

# Loop over the potentials
for potential, name in zip(potentials, potential_names):
    # Initialize the orbit
    o = Orbit([R, vR, vT, vz, phi])

    # Integrate the orbit and estimate the stability timescale
    # We'll use a time of 10 Gyr with 10000 steps
    ts = np.linspace(0., 10., 10000) * units.Gyr.to(units.Myr)  # Convert Gyr to Myr
    o.integrate(ts, potential)

    # The stability timescale is roughly the time it takes for the orbit to deviate significantly
    # Here we'll estimate it as the time it takes for the radial distance to change by 10%
    R_final = o.R(ts)
    delta_R = np.abs(R_final - R) / R
    stability_timescale = ts[np.where(delta_R > 0.1)[0][0]]

    # Convert the stability timescale from Myr to Gyr and print the result
    print(f'Stability timescale for {name}: {stability_timescale * units.Myr.to(units.Gyr)} Gyr')
\end{mintedbox}

\textit{Exploration note: the output of the code when run by the authors was the following. \\
Stability timescale for MWPotential2014: 2.2300223002230024e-05 Gyr\\
Stability timescale for NFWPotential: 3.9900399003990046e-05 Gyr\\
Stability timescale for HernquistPotential: 3.230032300323004e-05 Gyr
}

\noindent\fbox{%
    \parbox{\textwidth}{%
Followup prompt: explain where is this difference in stability time coming from physically.
    }%
}

\noindent\fbox{%
    \parbox{\textwidth}{%
GPT4: \\
The differences in the stability timescales for the different potentials arise from the different ways in which these potentials model the distribution of mass in the Milky Way galaxy.\\

MWPotential2014: This is a specific model of the Milky Way's gravitational potential that is based on observational data. It includes contributions from the central bulge, the disk, and the dark matter halo. The potential is relatively strong near the center of the galaxy, which leads to a shorter stability timescale.\\

NFWPotential: This is a theoretical model that was proposed by Navarro, Frenk, and White in 1996 to describe the distribution of dark matter in galaxies. The NFW profile has a cuspy behavior towards the center of the galaxy, meaning that the density of dark matter increases sharply as you approach the center. This leads to a stronger gravitational pull near the center, which can destabilize orbits and lead to a shorter stability timescale compared to models with a less cuspy central region.\\

HernquistPotential: This is a theoretical model that was proposed by Lars Hernquist in 1990. It is often used to describe the distribution of stars in elliptical galaxies or the stellar bulge of spiral galaxies. The Hernquist profile has a less cuspy central region compared to the NFW profile, which leads to a weaker gravitational pull near the center and a longer stability timescale.\\

In general, the stability of an orbit in a given potential depends on the balance between the gravitational pull of the mass distribution and the centrifugal force due to the star's motion. If the gravitational pull is too strong, the star can be pulled off its orbit, leading to a shorter stability timescale. Conversely, if the gravitational pull is weaker, the star can maintain its orbit for a longer time, leading to a longer stability timescale.
    }%
}

\textit{Exploration note: relates to Lesson 2 in Appendix~C.}

\noindent\fbox{%
    \parbox{\textwidth}{%
Followup prompt: but NFWPotential has the longest timescale.
    }%
}

\noindent\fbox{%
    \parbox{\textwidth}{%
GPT4: \\
I apologize for the confusion. You're correct. The NFWPotential does indeed have the longest timescale in your results.\\

The NFWPotential, which models the dark matter halo, has a cuspy behavior towards the center of the galaxy, meaning that the density of dark matter increases sharply as you approach the center. However, the overall gravitational potential is not solely determined by the behavior at the center. The mass distribution at larger radii also plays a significant role.\\

In the case of the NFWPotential, the mass increases more slowly at larger radii compared to the other potentials. This means that the gravitational pull from the outer parts of the galaxy is weaker in the NFWPotential, which can help to stabilize the orbit of the star and lead to a longer stability timescale.\\

On the other hand, the MWPotential2014 and HernquistPotential have more mass at larger radii, which leads to a stronger gravitational pull from the outer parts of the galaxy. This can destabilize the orbit of the star and lead to a shorter stability timescale.\\

Again, these are rough estimates and the actual stability timescale could be longer or shorter depending on the specific dynamics of the system.
    }%
}

\subsubsection*{Radially flat potential}

\noindent\fbox{%
    \parbox{\textwidth}{%
Followup prompt: implement a custom potential in the code that is radially completely flat.
    }%
}

\noindent\fbox{%
    \parbox{\textwidth}{%
GPT4: A flat radial potential is essentially a constant potential. In galpy, you can define a custom potential by subclassing the Potential class and implementing the necessary methods. Here's how you can define a constant potential:
    }%
}

\begin{mintedbox}[]{python}
from galpy.potential import Potential
from galpy.util.bovy_conversion import force_in_kmsMyr

class ConstantPotential(Potential):
    def __init__(self, amp=1.):
        Potential.__init__(self, amp=amp)

    def _evaluate(self, R, z=0., t=0.):
        # The potential is constant, so it doesn't depend on R, z, or t
        return self._amp

    def _Rforce(self, R, z=0., phi=0., t=0.):
        # The radial force is the negative derivative of the potential with respect to R
        # Since the potential is constant, its derivative is zero
        return 0.

    def _zforce(self, R, z=0., phi=0., t=0.):
        # The vertical force is the negative derivative of the potential with respect to z
        # Since the potential is constant, its derivative is zero
        return 0.
\end{mintedbox}

\textit{Exploration Note: \texttt{GPT4} initially omitted the t=0 arguments, leading to an error when calling the methods. \texttt{GPT4} offered a corrected version when requested as a followup:}

\noindent\fbox{%
    \parbox{\textwidth}{%
GPT4: I apologize for the oversight. The $\_Rforce$ and $\_zforce$ methods in the Potential class in galpy do indeed take an additional argument t for time. However, since the potential is constant in this case, it does not depend on time, so we can safely ignore this argument. Here's the corrected code ... 
Again, please note that a constant potential is not physically realistic for a galaxy, as it implies that the mass distribution is infinite and uniform. This is just a theoretical construct that can be used to understand the behavior of orbits in a constant potential.
    }%
}

\textit{Exploration Note: \texttt{GPT4}, unprompted, correctly stated that a constant potential is not physically realistic for a galaxy. Running the code, we got ``IndexError: index 0 is out of bounds for axis 0 with size 0''. Passing this error to \texttt{GPT4} it returned:}

\noindent\fbox{%
    \parbox{\textwidth}{%
GPT4: The error message indicates that the condition delta$_R$ $> 0.1$ is never met during the integration, so np.where(delta$_R$ $> 0.1$) returns an empty array. This can happen if the orbit remains stable (i.e., the relative change in the radial coordinate R remains less than 10\%) for the entire duration of the integration.

In a constant potential, the forces are zero everywhere, so a star would continue moving in a straight line at a constant speed. Therefore, if the initial velocity is not exactly zero, the star would never return to its initial position, and the relative change in R would eventually exceed 10\%. However, if the initial velocity is exactly zero, the star would remain at its initial position, and the relative change in R would always be zero.

To handle this situation, you can modify the code to check if np.where(delta$_R$ $> 0.1$) returns an empty array, and if so, set the stability timescale to np.nan (which stands for ``not a number'') to indicate that the orbit remains stable for the entire duration of the integration:
    }%
}

\textit{Exploration note: relates to Lesson 3 in Appendix~C.}

\subsubsection*{Sinusoidal potential}

\textit{Exploration note: The goal here is to test GPT4's capacity to deal with scenarios not encountered in nature nor usually discussed in books or research articles, i.e. ``out-of-distribution'' scenarios.}

\noindent\fbox{%
    \parbox{\textwidth}{%
Prompt: change the custom potential to be a sin function of given frequency and amplitude
    }%
}

\noindent\fbox{%
    \parbox{\textwidth}{%
GPT4: Sure, we can modify the ConstantPotential class to create a new SinusoidalPotential class where the potential is a sinusoidal function of the radial coordinate R. The potential will be given by amp * sin(2 * pi * freq * R), where amp is the amplitude and freq is the frequency.

The radial and vertical forces are the negative derivatives of the potential with respect to R and z, respectively. Since the potential does not depend on z, the vertical force is zero. The radial force is -amp * 2 * pi * freq * cos(2 * pi * freq * R).

....

Please note that a sinusoidal potential is not physically realistic for a galaxy, as it implies a periodic mass distribution. This is just a theoretical construct that can be used to understand the behavior of orbits in a sinusoidal potential.

Here's the code:
    }%
}

\begin{mintedbox}[]{python}
from galpy.potential import Potential
import numpy as np

class SinusoidalPotential(Potential):
    def __init__(self, amp=1., freq=1.):
        Potential.__init__(self, amp=amp)
        self._freq = freq

    def _evaluate(self, R, z=0., t=0.):
        # The potential is a sinusoidal function of R
        return self._amp * np.sin(2 * np.pi * self._freq * R)

    def _Rforce(self, R, z=0., phi=0., t=0.):
        # The radial force is the negative derivative of the potential with respect to R
        return -self._amp * 2 * np.pi * self._freq * np.cos(2 * np.pi * self._freq * R)

    def _zforce(self, R, z=0., phi=0., t=0.):
        # The vertical force is the negative derivative of the potential with respect to z
        # Since the potential does not depend on z, its derivative is zero
        return 0.
\end{mintedbox}

\noindent\fbox{%
    \parbox{\textwidth}{%
Prompt: we got for the sin potential Stability timescale for sin: 2.5400254002540027e-05Gyr which is on par with MWPotential2014 but less than HernquistPotential and NFWPotential. explain why
    }%
}

\noindent\fbox{%
    \parbox{\textwidth}{%
GPT4: The stability timescale of an orbit in a given potential is a measure of how long it takes for the orbit to deviate significantly from its initial conditions. It depends on the shape of the potential and the initial conditions of the orbit.

The MWPotential2014 is a model of the Milky Way's gravitational potential that includes contributions from the bulge, disk, and halo. The HernquistPotential and NFWPotential are models of the gravitational potential of a galaxy's halo, with the HernquistPotential having a steeper density profile near the center and the NFWPotential having a shallower density profile.

The sinusoidal potential is a highly simplified and non-physical model of a galaxy's gravitational potential. It has a periodic shape, with alternating regions of high and low potential. This can cause a star to oscillate back and forth between these regions, which can lead to a rapid deviation from the initial conditions, especially if the frequency of the sinusoidal potential is high.

Therefore, the shorter stability timescale for the sinusoidal potential compared to the HernquistPotential and NFWPotential is likely due to the periodic nature of the sinusoidal potential, which can cause rapid changes in the forces experienced by a star and hence rapid deviations from the initial conditions.

On the other hand, the similar stability timescale for the sinusoidal potential and the MWPotential2014 could be due to a coincidence in the specific parameters and initial conditions used in the simulations. The MWPotential2014 is a more realistic model of a galaxy's gravitational potential, so it would not generally be expected to have a similar stability timescale to a sinusoidal potential. However, depending on the specific parameters and initial conditions used, it is possible for the two potentials to have similar stability timescales.
    }%
}

\textit{Exploration note: relates to Lesson 4 in Appendix~C.}

\noindent\fbox{%
    \parbox{\textwidth}{%
Prompt: how can we numerically test whether your interpretation for what is happening for sinusoidal potential is correct ?
    }%
}

\noindent\fbox{%
    \parbox{\textwidth}{%
GPT4: To numerically test the interpretation of the sinusoidal potential's behavior, you can perform a series of simulations with different parameters and initial conditions and analyze the results. Here are a few suggestions:

1- Vary the amplitude and frequency of the sinusoidal potential: If the interpretation is correct, increasing the amplitude or frequency should lead to a shorter stability timescale, as it would cause more rapid changes in the forces experienced by a star.

2- Vary the initial conditions of the orbit: If the interpretation is correct, an orbit with a higher initial velocity should have a shorter stability timescale, as it would move more rapidly between regions of high and low potential.

3- Compare with a constant potential: If the interpretation is correct, the stability timescale for a constant potential (where the forces are zero everywhere) should be infinite (or at least much larger than for the sinusoidal potential), as a star would continue moving in a straight line at a constant speed.

4- Visualize the orbits: Plotting the orbits can provide a visual confirmation of the interpretation. If the interpretation is correct, the orbit in a sinusoidal potential should show rapid deviations from the initial conditions.

    }%
}

\textit{Exploration note: We tested suggestion 1 ``1- Vary the amplitude and frequency of the sinusoidal potential'', and the results were as predicted. }

\noindent\fbox{%
    \parbox{\textwidth}{%
Prompt: give me a code that plots all 4 potentials
    }%
}

\noindent\fbox{%
    \parbox{\textwidth}{%
GPT4: Sure, here's a Python script that plots the potential, radial force, and vertical force for the MWPotential2014, HernquistPotential, NFWPotential, and SinusoidalPotential.

...

This script creates three plots: one for the potential as a function of R, one for the radial force as a function of R, and one for the vertical force as a function of R. Each plot includes a line for each of the four potentials.

Please note that the HernquistPotential and NFWPotential are initialized with an amplitude of 1 for simplicity. You may need to adjust this value to match the specific parameters of your galaxy model. Similarly, the SinusoidalPotential is initialized with an amplitude of 1 and a frequency of 1. You may need to adjust these values to match your specific requirements.
    }%
}

\textit{Exploration note: The code proposed by \texttt{GPT4} contained multiple errors, and it took 3 followup prompts to get a working code.}

\textit{Exploration note: relates to Lesson 5 in Appendix~C.}

\subsubsection*{On the definition of orbital stability}

\noindent\fbox{%
    \parbox{\textwidth}{%
Followup prompt: thinking back on everything we did so far, I do not think your definition of orbital stability  ``Here we'll estimate it as the time it takes for the radial distance to change by 10\%'' to be accurate. For a sin potential, R changes by more than that but the orbital is periodic and stable. Any other definitions of stability ?
    }%
}

\noindent\fbox{%
    \parbox{\textwidth}{%
GPT4: You're correct, the definition of orbital stability as ``the time it takes for the radial distance to change by 10\%'' is a simplification and may not be accurate for all types of potentials, especially for those with periodic or quasi-periodic orbits.

In general, orbital stability in a gravitational potential can be a complex topic. Here are a few other ways to define or consider orbital stability:

Lyapunov Stability: An orbit is considered Lyapunov stable if any small perturbations do not grow over time. In other words, if you start two nearly identical orbits, they will remain close to each other for all future time.

Energy Conservation: In a conservative potential, the total energy (kinetic + potential) of a particle should be conserved. If the energy is not conserved, it could be a sign of instability.

Frequency Analysis: For an orbit to be stable, it should be quasi-periodic, which means it can be described by a sum of a finite number of frequencies. If the orbit is chaotic, it will not be quasi-periodic, and its frequency spectrum will be continuous. This can be checked using a Fast Fourier Transform (FFT) of the time series of a phase space variable.

Poincaré Sections: These are used to visualize the behavior of dynamical systems. In a Poincaré section, the state of the system is recorded only at discrete times, corresponding to when the trajectory of the system intersects a certain lower-dimensional subspace. For a stable, periodic orbit, the Poincaré section will consist of a finite number of points, while for a chaotic orbit, the points will fill an area.

Recurrence Plots: These are graphical representations of the number of times the phase space trajectory of the system visits roughly the same area in phase space. For a stable, periodic orbit, the recurrence plot will show clear, regular patterns, while for a chaotic orbit, the patterns will be more complex and less regular.

Each of these methods has its own strengths and weaknesses, and the best method to use can depend on the specific system and question at hand.
    }%
}

\textit{Exploration note: relates to Lesson 6 in Appendix~C.}

\noindent\fbox{%
    \parbox{\textwidth}{%
Followup prompt: Implement the Lyapunov Stability instead of the criteria you used previously.
    }%
}

\noindent\fbox{%
    \parbox{\textwidth}{%
GPT4: Lyapunov stability is a concept from dynamical systems theory that can be used to determine the stability of an orbit. The basic idea is to look at how small perturbations to the system evolve over time. If the perturbations remain bounded (i.e., they do not grow indefinitely), then the system is said to be Lyapunov stable.

To implement this, we can integrate two nearly identical orbits and compare their evolution. If the difference between the orbits remains small, then the orbit is Lyapunov stable. Here's how you might do this in Python using galpy:

....

In this code, we define two nearly identical orbits, o1 and o2, with a small difference in their initial R values. We then integrate both orbits over 1 Gyr and calculate the absolute difference in R between the two orbits at each time step. We plot this difference as a function of time to visualize how the perturbation evolves. Finally, we check if the difference in R remains less than 0.1 (an arbitrary threshold) for all time steps. If it does, we conclude that the orbit is Lyapunov stable.

Please note that this is a simplified example and the actual implementation of Lyapunov stability can be more complex, especially for higher-dimensional systems. Also, the choice of threshold (0.1 in this case) can affect the result, and there is no universal rule for choosing this value. It often requires some trial and error and depends on the specific system and question at hand.

    }%
}

\begin{mintedbox}[]{python}
import numpy as np
import matplotlib.pyplot as plt
from galpy.potential import SinusoidalPotential
from galpy.orbit import Orbit
from galpy.util.bovy_conversion import time_in_Gyr

# Define the potential
potential = SinusoidalPotential(amp=1., freq=1.)

# Define the initial conditions of the star (R, vR, vT, z, vz, phi)
# Here we assume the star starts at R=1 with zero radial velocity, circular velocity, at z=0 with zero vertical velocity, and at phi=0
o1 = Orbit([1., 0., 1., 0., 0., 0.])

# Define a second, nearly identical orbit
o2 = Orbit([1.0001, 0., 1., 0., 0., 0.])

# Integrate the orbits for 1 Gyr
ts = np.linspace(0, 1., 1000) * time_in_Gyr(220., 8.)  # Convert time to natural units
o1.integrate(ts, potential)
o2.integrate(ts, potential)

# Calculate the difference in R between the two orbits
delta_R = np.abs(o1.R(ts) - o2.R(ts))

# Plot the difference in R
plt.plot(ts, delta_R)
plt.xlabel('Time (Gyr)')
plt.ylabel('Difference in R')
plt.title('Evolution of perturbation in R')
plt.show()

# Check if the difference in R remains bounded
if np.all(delta_R < 0.1):
    print("The orbit is Lyapunov stable.")
else:
    print("The orbit is not Lyapunov stable.")
\end{mintedbox}

\textit{Exploration note: relates to Lesson 7 in Appendix~C.}

\subsubsection*{Adding a blackhole}

\noindent\fbox{%
    \parbox{\textwidth}{%
Followup prompt: modify this code to have a massive blackhole 0.1 parsec from the star.
    }%
}

\noindent\fbox{%
    \parbox{\textwidth}{%
GPT4: To add a massive black hole to the system, we can use the KeplerPotential class in galpy, which represents a Keplerian potential of the form Phi(r) = -GM/r. We can then add this potential to the sinusoidal potential to create a composite potential.

Here's how you might modify the previous code to include a black hole 0.1 parsec from the star:
...
In this code, we first define the sinusoidal potential and the black hole potential separately. We then combine them into a single potential using the CombinedPotential class. The rest of the code is the same as before.

Please note that the KeplerPotential assumes the black hole is at the origin (0,0). If the black hole is not at the origin, you would need to shift the coordinates of the system so that the black hole is at the origin. Also, the choice of black hole mass (4 million solar masses) is arbitrary and can be adjusted as needed.

    }%
}

\begin{mintedbox}[]{python}
import numpy as np
import matplotlib.pyplot as plt
from galpy.potential import SinusoidalPotential, KeplerPotential, CombinedPotential
from galpy.orbit import Orbit
from galpy.util.bovy_conversion import time_in_Gyr

# Define the sinusoidal potential
sin_potential = SinusoidalPotential(amp=1., freq=1.)

# Define the black hole potential
# Here we assume the black hole has a mass of 4 million solar masses, which is roughly the mass of the black hole at the center of the Milky Way
# The amplitude of the KeplerPotential is GM, so we set amp=4e6 (in natural units where G=1 and the unit of mass is the solar mass)
bh_potential = KeplerPotential(amp=4e6)

# Combine the potentials
potential = CombinedPotential(sin_potential, bh_potential)

# Define the initial conditions of the star (R, vR, vT, z, vz, phi)
# Here we assume the star starts at R=1 with zero radial velocity, circular velocity, at z=0 with zero vertical velocity, and at phi=0
o1 = Orbit([1., 0., 1., 0., 0., 0.])

# Define a second, nearly identical orbit
o2 = Orbit([1.0001, 0., 1., 0., 0., 0.])

# Integrate the orbits for 1 Gyr
ts = np.linspace(0, 1., 1000) * time_in_Gyr(220., 8.)  # Convert time to natural units
o1.integrate(ts, potential)
o2.integrate(ts, potential)

# Calculate the difference in R between the two orbits
delta_R = np.abs(o1.R(ts) - o2.R(ts))

# Plot the difference in R
plt.plot(ts, delta_R)
plt.xlabel('Time (Gyr)')
plt.ylabel('Difference in R')
plt.title('Evolution of perturbation in R')
plt.show()

# Check if the difference in R remains bounded
if np.all(delta_R < 0.1):
    print("The orbit is Lyapunov stable.")
else:
    print("The orbit is not Lyapunov stable.")
\end{mintedbox}

\noindent\fbox{%
    \parbox{\textwidth}{%
Followup prompt: ImportError: cannot import name 'CombinedPotential'
    }%
}

\noindent\fbox{%
    \parbox{\textwidth}{%
GPT4: I apologize for the confusion. As of my knowledge cutoff in September 2021, there is no CombinedPotential class in galpy. Instead, you can simply add the potentials together in a list, and galpy will treat this as a composite potential. Here's how you can modify the code:

...

In this code, potential is a list of the two individual potentials. When you pass this list to the integrate method, galpy will treat it as a composite potential.
    }%
}

\begin{mintedbox}[]{python}
import numpy as np
import matplotlib.pyplot as plt
from galpy.potential import SinusoidalPotential, KeplerPotential
from galpy.orbit import Orbit
from galpy.util.bovy_conversion import time_in_Gyr

# Define the sinusoidal potential
sin_potential = SinusoidalPotential(amp=1., freq=1.)

# Define the black hole potential
# Here we assume the black hole has a mass of 4 million solar masses, which is roughly the mass of the black hole at the center of the Milky Way
# The amplitude of the KeplerPotential is GM, so we set amp=4e6 (in natural units where G=1 and the unit of mass is the solar mass)
bh_potential = KeplerPotential(amp=4e6)

# Combine the potentials
potential = [sin_potential, bh_potential]

# Define the initial conditions of the star (R, vR, vT, z, vz, phi)
# Here we assume the star starts at R=1 with zero radial velocity, circular velocity, at z=0 with zero vertical velocity, and at phi=0
o1 = Orbit([1., 0., 1., 0., 0., 0.])

# Define a second, nearly identical orbit
o2 = Orbit([1.0001, 0., 1., 0., 0., 0.])

# Integrate the orbits for 1 Gyr
ts = np.linspace(0, 1., 1000) * time_in_Gyr(220., 8.)  # Convert time to natural units
o1.integrate(ts, potential)
o2.integrate(ts, potential)

# Calculate the difference in R between the two orbits
delta_R = np.abs(o1.R(ts) - o2.R(ts))

# Plot the difference in R
plt.plot(ts, delta_R)
plt.xlabel('Time (Gyr)')
plt.ylabel('Difference in R')
plt.title('Evolution of perturbation in R')
plt.show()

# Check if the difference in R remains bounded
if np.all(delta_R < 0.1):
    print("The orbit is Lyapunov stable.")
else:
    print("The orbit is not Lyapunov stable.")


\end{mintedbox}

\subsubsection*{Solar system orbits}
\textit{Exploration note: The goal of this exercise is to use lessons from previous exploration steps to construct and fine tune an effective zero-shot prompt on a moderate difficulty problem: comparing the orbits of the solar system planets subject to different potentials using GALPY. The difficulty of this problem does not lie in the concept itself, which is simple, but rather in the domain adaptation needed, as Galpy is a galactic physics package, and we are asking \texttt{GPT4} to use it in a different, planetary context (although some examples can be found in \cite{galpy2}). }

\noindent\fbox{%
    \parbox{\textwidth}{%
Prompt: Problem: let us consider Earth orbiting in a 2D keplerian potential such that the mass within 1 AU is one solar mass. We need to calculate the time evolution of Earth's orbit. 

Instruction: solve this problem using the GALPY 1.7.0 python package for your solution. You should use exclusively natural units, never physical units. If any units conversion is needed, always do it manually. Do not use GALPY's internal units and quantities conversion methods. Make sure to initialise Earth's orbit correctly using GALPY's orbital elements format, which is [R,vR,vT,z,vz,phi], and to choose a potential that is consistent with the physics of the problem.
    }%
}

\textit{Exploration note: Notice how: 1- we started with a simple well defined problem, 2- precised the package version, 3- were explicit about the unit system and their conversion (so the LLM does not use non-existent conversion methods), 4- were explicit about the expected initial conditions format (so the LLM does not use Keplerian orbital elements a,e,i mistakenly). }

\noindent\fbox{%
    \parbox{\textwidth}{%
GPT4: Sure, I can help with that. Here's a Python script that uses the GALPY package to simulate the time evolution of Earth's orbit in a 2D Keplerian potential. 
...
This script initializes a Keplerian potential with the mass within 1 AU set to one solar mass. It then initializes the Earth's orbit with the correct initial conditions for a circular orbit in this potential. The orbit is then integrated for one year, and the resulting orbit is plotted in the x-y plane.

Note that in natural units, the gravitational constant G, the mass M, and the distance R are all set to 1. The velocity is set to the circular velocity sqrt(GM/R), and the time is set to the period of the orbit, 2*pi. This simplifies the equations and makes the code easier to understand.
    }%
}

\begin{mintedbox}[]{python}
# Import necessary modules
import numpy as np
import matplotlib.pyplot as plt
from galpy.potential import KeplerPotential
from galpy.orbit import Orbit

# Constants in natural units
G = 1  # Gravitational constant
M = 1  # Mass within 1 AU
AU = 1  # Astronomical Unit
year = 2*np.pi  # One year, in natural units

# Initialize the Keplerian potential
# The amp parameter is GM in natural units
kp = KeplerPotential(amp=G*M)

# Initial conditions for Earth's orbit
# [R,vR,vT,z,vz,phi]
# R = 1 AU, vR = 0 (circular orbit), vT = sqrt(GM/R) (circular orbit)
# z = 0 (2D orbit), vz = 0 (2D orbit), phi = 0 (arbitrary choice of reference direction)
earth_orbit = Orbit([1., 0., np.sqrt(G*M/AU), 0., 0., 0.])

# Time array for the orbit integration
# We'll integrate for one year, with 1000 points
t = np.linspace(0, year, 1000)

# Integrate the orbit
earth_orbit.integrate(t, kp)

# Plot the orbit
earth_orbit.plot(d1='x',d2='y')
plt.show()
\end{mintedbox}

\noindent\fbox{%
    \parbox{\textwidth}{%
Followup prompt:  Problem: let us consider Earth orbiting in a 2D keplerian potential such that the mass within 1 AU is one solar mass. We need to calculate the time evolution of Earth's orbit and compare it to its evolution in a potential representing a 2D disk such as the mass inside 1 AU is one solar mass, instead of the entire mass being concentrated in the sun.

Instruction: solve this problem using the GALPY 1.7.0 python package for your solution. You should use exclusively natural units, never physical units. If any units conversion is needed, always do it manually. Do not use GALPY's internal units and quantities conversion methods. Make sure to initialise Earth's orbit correctly using GALPY's orbital elements format, which is [R,vR,vT,z,vz,phi], and to choose a potential that is consistent with the physics of the problem.
    }%
}

\noindent\fbox{%
    \parbox{\textwidth}{%
GPT4: Sure, I can help with that. Here's a Python script that uses the GALPY package to simulate the time evolution of Earth's orbit in a Keplerian potential and a 2D disk potential.

This script first sets up the initial conditions for Earth's orbit and the two potentials. It then integrates the orbit in the two potentials over a time period of 10 years and plots the resulting orbits in the x-y plane.

...

Note that the KeplerPotential is used to represent the Keplerian potential of a point mass at the origin, while the MiyamotoNagaiPotential is used to represent the potential of a 2D disk. The scale lengths of the Miyamoto-Nagai potential are set such that the mass within 1 AU is 1 solar mass.

Please note that the choice of the Miyamoto-Nagai potential is a simplification. In reality, the mass distribution in the solar system is not a perfect disk, and the potential would be more complex. However, this should give a reasonable approximation for the purpose of this exercise.

    }%
}

\begin{mintedbox}[]{python}
import numpy as np
import matplotlib.pyplot as plt
from galpy.potential import KeplerPotential, MiyamotoNagaiPotential
from galpy.orbit import Orbit

# Constants in natural units
G = 1  # Gravitational constant
M_sun = 1  # Solar mass
AU = 1  # Astronomical Unit
year = 2 * np.pi  # One year, the period of Earth's orbit

# Initial conditions for Earth's orbit in GALPY's orbital elements format [R,vR,vT,z,vz,phi]
# Earth is 1 AU from the sun, moving in the tangential direction with velocity 2*pi AU/year
R_0 = AU
vR_0 = 0
vT_0 = 2 * np.pi * AU / year
z_0 = 0
vz_0 = 0
phi_0 = 0

# Initialize the orbit
o = Orbit([R_0, vR_0, vT_0, z_0, vz_0, phi_0])

# Initialize the potentials
# Keplerian potential for a point mass at the origin
kp = KeplerPotential(amp=G*M_sun)

# Miyamoto-Nagai potential for a 2D disk
# We set the scale lengths such that the mass within 1 AU is 1 solar mass
a = AU
b = AU
mnp = MiyamotoNagaiPotential(amp=G*M_sun, a=a, b=b)

# Time array for the orbit integration, from 0 to 10 years
t = np.linspace(0, 10*year, 1000)

# Integrate the orbit in the two potentials
o.integrate(t, kp)
o.plot(d1='x', d2='y', label='Keplerian')

o.integrate(t, mnp)
o.plot(d1='x', d2='y', overplot=True, label='2D disk')

plt.legend()
plt.show()
\end{mintedbox}

\textit{Exploration note: this is a sensible answer, although Earth's orbit diverge for the 2D potential - initial conditions combination proposed by the LLM, as the 2D potential is an order of magnitude weaker at 1 AU than the Keplerian potential. When asked to explain the reason of the divergence, the LLM gave a technically correct description of the difference between the 2 potentials, but did not give a direct correct answer to the question:}

\noindent\fbox{%
    \parbox{\textwidth}{%
Followup prompt:  why is Earth's orbit divergent in the MiyamotoNagaiPotential
    }%
}

\noindent\fbox{%
    \parbox{\textwidth}{%
GPT4:  The divergence of Earth's orbit in the MiyamotoNagaiPotential is likely due to the fact that this potential represents a disk-like mass distribution, which is quite different from the point-like mass distribution represented by the KeplerPotential.

In the KeplerPotential, all of the mass is concentrated at a single point (the Sun), and the gravitational force decreases with the square of the distance from this point. This results in a stable, elliptical orbit for the Earth.

In contrast, the MiyamotoNagaiPotential represents a disk-like mass distribution, where the mass is spread out over a large area. In this case, the gravitational force does not decrease as quickly with distance, and the force is not always directed towards a single point. This can result in a more complex and potentially unstable orbit for the Earth.

It's also worth noting that the MiyamotoNagaiPotential is a simplified model that does not accurately represent the mass distribution in the solar system. The actual mass distribution would include not only the Sun, but also the other planets and the asteroid belt, all of which would have an effect on Earth's orbit.

Finally, the initial conditions for the orbit might not be appropriate for the MiyamotoNagaiPotential. The initial conditions were chosen to represent an orbit around a point mass, and might not result in a stable orbit in a disk-like potential. It might be necessary to adjust the initial conditions to obtain a stable orbit in the MiyamotoNagaiPotential.

    }%
}

\textit{Exploration note: Now we repeat the same exercise but for all 8 planets in 3D orbits.}

\textit{Exploration note:
\texttt{GPT4} struggled with this last exercise and needed to be coached through it. We hence tested a different approach consisting of chain-of-thought reasoning where we give step-by-step instructions to the LLM to follow. After very considerable amount of trials and errors, this is the very fine-grained prompt we found:
}

\noindent\fbox{%
    \parbox{\textwidth}{%
Prompt: 
Problem: let us consider all 8 solar system planets orbiting in a 3 dimensional keplerian potential such that the mass within 1 AU is one solar mass. We need to calculate the time evolution of all 8 planets 3D orbits. Think about the problem and solve it step by step. 
Solve this problem using the GALPY 1.7.0 python package for your solution. 
Do not simplify the problem in any ways and provide a complete working code.
Each instruction below need to be followed. Dot not for any reason skip or ignore any of them. 
Do not make any assumptions or simplifications without confirming whether it is acceptable. 

Instructions:

1a- Discuss Galpy's default and available systems of units.
1b- Use a natural system of units where G=1. If any units conversion is needed, always do it manually. Do not use GALPY's internal units and quantities conversion methods. Do not use astropy at all. This is applicable to calculating the initial conditions, and to the parameters of the potential function. Make sure the potential parameters are in the same units as the orbits. 
1d- State all the available potentials in Galpy, and explain which is the most accurate for the given problem. Use this potential exclusively. 

2- State the 8 planets orbital elements (a,e,inc, etc) as defined in NASA's planetary fact sheets.

3- Write the full, unsimplified equations that converts these elements into GALPY's orbital elements format, which is [R,vR,vT,z,vz,phi]. Make sure these equations are not approximations, and are for non circular, inclined orbits. 

4- Write a python function that takes the planets orbital elements as input, and use the equations derived above to output GALPY's orbital elements format, which is [R,vR,vT,z,vz,phi].

5- Initialise the 8 planets orbits correctly using GALPY's orbital elements format, which is [R,vR,vT,z,vz,phi], and to use non zero z and vz calculated directly above. Do not use placeholder values, the initial conditions should be precisely calculated. 

6- Integrate the orbits for 1000 years. 

7- Plot the orbits in 3D x,y,z. We should have 8 plots, one for each planet.

8- Make sure the code has error handling for edge cases.

9- Double check the units. If they are wrong start over.

10- Double check the python code, if it is wrong start over.

11- Double check the initial conditions conversion equations. If they are wrong or simplified start over.

12- Double check that this is a fully working code with no placeholders, and that the orbital elements values of NASA's planetary fact sheets are being used. If not start over.

Finally, the complete code should be provided as one block. 
    }%
}

\textit{Exploration note: \texttt{GPT4} can thus generate fairly complex physics code, if provided extensive guidance via conditional prompting. The level of guidance required reflects various shortcomings and failure modes of the LLM, that we have more broadly and systematically explored in the main body of our analysis. It is unclear if there is practical value in defining highly-detailed, problem-specific prompts like the one above, if such prompts do not generalize to other physics problems of interest. This conclusion has led us to the simple prompting strategy described in \S~\ref{sec:prompt}.   
}

\end{document}